\documentclass[10pt,twocolumn,letterpaper]{article}

\usepackage[pagenumbers]{cvpr}

\usepackage{graphicx}
\usepackage{amsmath}
\usepackage{amssymb}
\usepackage{booktabs}
\usepackage{url}
\usepackage{amsfonts}
\usepackage{multirow}
\usepackage{tabularx}
\usepackage{color}
\usepackage{colortbl}
\usepackage{epsfig}
\usepackage{mathtools}
\usepackage{courier}
\usepackage{makecell}
\usepackage{float}
\usepackage{bbm}
\usepackage{pifont}
\usepackage{breakcites}
\usepackage{savesym}
\usepackage{wrapfig}
\usepackage{dsfont}
\usepackage{listings}
\usepackage{mdframed}
\usepackage[dvipsnames]{xcolor}
\usepackage[hang,flushmargin,symbol]{footmisc} 
\definecolor{darkgreen}{HTML}{20cb07}
\newcommand{\red}[1]{{\color{red}#1}}
\newcommand{\cmark}{\color{darkgreen}\ding{51}\color{black}}
\newcommand{\xmark}{\color{red}\ding{55}\color{black}}

\newcommand{\oursjcef}[0]{JCEF}
\newcommand{\fulljcef}[0]{Just Caption Every Frame (\oursjcef)}

\newcommand{\jcef}[0]{\oursjcef{}}
\newcommand{\jceffull}[0]{\fulljcef{}}

\newcommand{\ours}[0]{MoReVQA}
\newcommand{\oursfull}[0]{Modular Reasoning for Video Question Answering (\ours{})}

\newcommand{\para}[1]{\noindent\textbf{#1}~}

\newcommand{\tightset}[1]{\{#1\}}
\newcommand{\deemph}[1]{\color{gray}#1}

\newcommand*{\affmark}[1][*]{\textsuperscript{#1}}

\definecolor{cvprblue}{rgb}{0.21,0.49,0.74}
\usepackage[pagebackref,breaklinks,colorlinks,citecolor=cvprblue]{hyperref}

\title{MoReVQA: Exploring Modular Reasoning Models for\\Video Question Answering
}

\author{Juhong Min\affmark[1,2]\footnotemark[1]\hspace{0.8cm}Shyamal Buch\affmark[1]\hspace{0.8cm}Arsha Nagrani\affmark[1]\hspace{0.8cm}Minsu Cho\affmark[2]\hspace{0.8cm}Cordelia Schmid\affmark[1]\vspace{1.5mm}\\
\affmark[1]Google\hspace{1.5cm} 
\affmark[2]POSTECH\footnotemark[2] \\
{\tt\small \url{http://juhongm999.github.io/morevqa}}
\vspace{-0.5cm} 
}

\begin{document}
\maketitle
\begin{abstract}
This paper addresses the task of video question answering (videoQA) via a decomposed multi-stage, modular reasoning framework. Previous modular methods have shown promise with a single planning stage ungrounded in visual content. However, through a simple and effective baseline, we find that such systems can lead to brittle behavior in practice for challenging videoQA settings. Thus, unlike traditional single-stage planning methods, we propose a multi-stage system consisting of an event parser, a grounding stage, and a final reasoning stage in conjunction with an external memory. All stages are training-free, and performed using few-shot prompting of large models, creating interpretable intermediate outputs at each stage. By decomposing the underlying planning and task complexity, our method, \ours{}, improves over prior work on standard videoQA benchmarks (NExT-QA, iVQA, EgoSchema, ActivityNet-QA) with state-of-the-art results, and extensions to related tasks (grounded videoQA, paragraph captioning).
\end{abstract}

\footnotetext[1]{Work done while Student Researcher intern at Google.}
\footnotetext[2]{Pohang University of Technology and Science}

\section{Introduction}
\label{sec:introduction}

\begin{figure}[t]
    \begin{center}
        \includegraphics[width=1.00\linewidth]{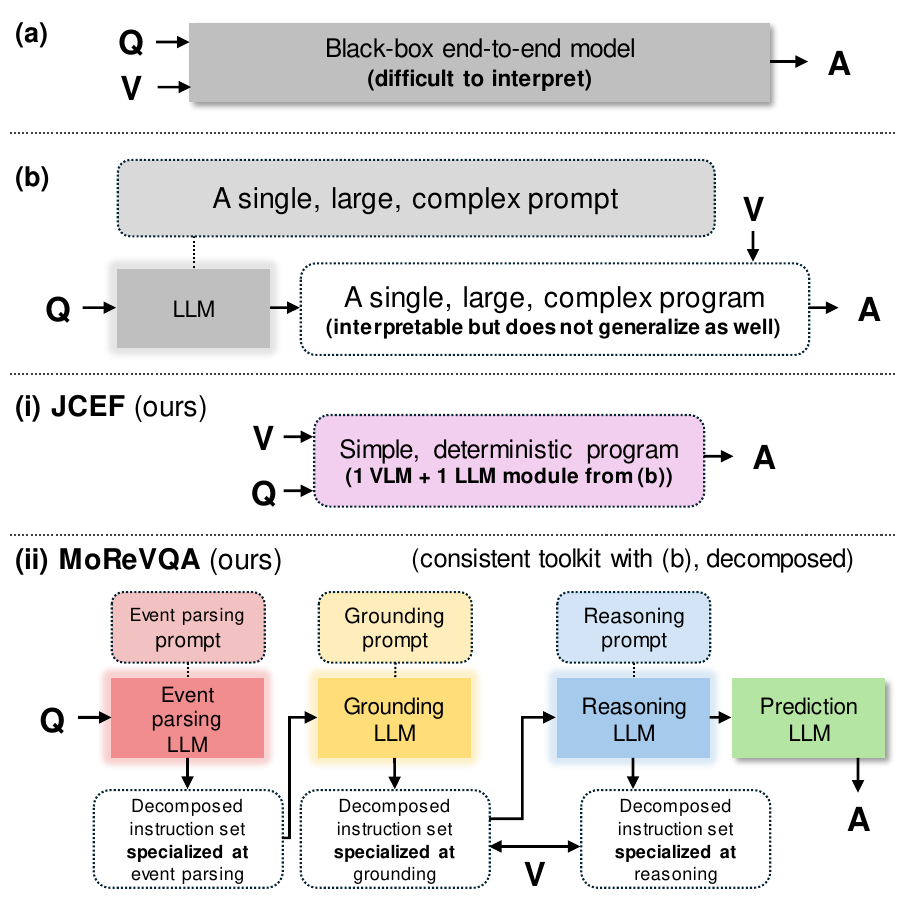}
    \end{center}
    \vspace{-3.0mm} 
      \caption{\textbf{\ours: a new multi-stage, modular reasoning model for videoQA}.
      Prior work relies on either (a)~black-box end-to-end models that are difficult to interpret, or (b)~modular systems where an interpretable planning step (program generation) is done in a single, ungrounded stage. (i) In this work, we find that single-stage planning leads in practice to brittle behavior, underperforming a new simple baseline (JCEF) that captions frames and predicts an answer (with two modules from (b)). (ii) We then introduce our new \ours{} method incorporating both \textit{modularity} and \textit{multi-stage planning}, providing interpretable, grounded planning and execution traces, while simultaneously delivering improvements in overall accuracy by effectively decomposing the underlying task complexity (still using consistent base models with (b)). Above: \textit{Q} is question, \textit{V} is video, \textit{A} is answer.}
    \vspace{-4.0mm} 
\label{fig:teaser}
\end{figure}

The predominant approach for solving video understanding tasks such as video question answering (videoQA) has long been end-to-end networks~\cite{alayrac2022flamingo,wang2022git,wang2021end,seo2022end,chen2022pali,chen2023pali}. A major challenge with such methods, however, is their black-box nature -- leading to a lack of interpretability and compositional generalization. For videos in particular, an important desired capability is the ability to understand events at different temporal scales, which is challenging for existing end-to-end vision-language models (VLMs) that typically see only a few frames~\cite{wang2022git,chen2022pali,chen2023pali}. This has led to a recent interest in modular or programmatic approaches~\cite{suris2023vipergpt,subramanian2023modular,gupta2023visual} to solve such problems, particularly leveraging the success of large language models (LLMs)~\cite{chowdhery2022palm,zhang2022opt,touvron2023llama} which have shown impressive reasoning and planning capabilities.
These methods generate symbolic programs~\cite{subramanian2023modular,suris2023vipergpt} using an LLM capable of producing code. They are interpretable and can be executed directly (leveraging independent visual or language processing modules). Their advantages are that they are training-free, compositional, and achieve impressive performance on few-shot vision and language tasks.

In this paper, we analyze the performance of such methods in closer detail, particularly for the case of videoQA (across 4 datasets \cite{xiao2021next,mangalam2023egoschema,yang2021just,yu2019activitynet}, representative of a range of video domains, lengths, and question types) and for single stage modular frameworks (such as ViperGPT~\cite{suris2023vipergpt}).
We find that, while recent modular approaches building on large, state-of-the-art end-to-end networks (LLMs and VLMs) as modules have shown significant promise~\cite{suris2023vipergpt}, a simple Socratic~\cite{zeng2022socratic} baseline, which we call \fulljcef, based on the \textit{same} underlying models, can actually \textit{outperform} these prior approaches by a significant margin.
As the name suggests, \oursjcef{} simply captions every frame in the video using a large vision-language model (VLM)~\cite{chen2023pali3}, and then feeds all captions along with the question to an LLM to produce an answer (Fig.~\ref{fig:teaser}(i) and Fig. \ref{fig:jcef}).
We hypothesize that the reason this baseline outperforms prior work is that these modular frameworks (Fig. \ref{fig:teaser}(b)) consist of a \textit{single planning stage} which may be ungrounded in the video (\ie the entire program or set of steps to be executed is determined in a single stage directly from the language prompt alone), and hence in practice the single-stage planner must be prompted with a large space of complex combinations required for answering diverse questions in video~\cite{khandelwal2023analyzing}.  While the performance of JCEF is impressive, it is less interpretable than the previously mentioned modular approaches, as captions for each frame tend to be generic and not question-specific (Fig. \ref{fig:jcef}).

In this work, we propose a decomposed, modular, and multi-stage approach for video question answering to address these limitations (Fig.~\ref{fig:teaser}(ii) and Fig.~\ref{fig:ours_modular_architecture}). Our method consists of three key planning and execution stages: (1) \textit{event parsing} that explicitly decomposes the events in the question, (2) \textit{grounding} that identifies corresponding temporal regions in the video that merit further tool use (so that every single frame does not have to be processed in detail), and (3) \textit{reasoning} that gives the final answer after considering the outputs of composed modules/APIs and the shared memory. This decomposition of single-stage planning is motivated by natural sub-tasks for videoQA and related video-language reasoning tasks.
All stages are training-free, and involve few-shot or zero-shot prompting of off-the-shelf modules (consistent with the API behavior in single-stage planning methods), in conjunction with an external read/write memory that maintains state and enables a more flexible design. We call our method \oursfull, and show that it outperforms \oursjcef~and other key single-stage modular baselines, while providing a grounded, interpretable planning and execution trace.

We summarize our key contributions as follows: \textbf{(1)}~We find that existing single-stage code-generation frameworks, while being modular and interpretable, are not necessarily well-suited for the complexity of generalizable VideoQA, and can be outperformed by a simple baseline we propose using a subset of its tool components (\eg a large VLM and LLM), \textbf{(2)} we design a multi-stage modular reasoning system (\ours) that alleviates this issue by decomposing the underlying planning sub-tasks effectively, and \textbf{(3)} we achieve state-of-the-art results on four standard videoQA benchmarks (NExT-QA, iVQA, EgoSchema, and ActivityNet-QA) across training-free (zero-shot/few-shot) methods, in some cases even outperforming fully-finetuned prior work. We also consider extensions to grounded videoQA (NExT-GQA \cite{xiao2023can}) and paragraph captioning (ActivityNet-Para \cite{krishna2017dense}) with strong performance.
\section{Related Work}
\label{sec:related_work}
\label{sec:related_work:end_to_end_models}
\label{sec:related_work:visual_programming}
\label{sec:related_work:socratic_models}

\para{VideoQA.} Video Question-Answering (videoQA) is a key task for multimodal video understanding systems to assess their ability to reason about a video~\cite{xu2017video,zhong2022video,xiao2021next,yang2021just,mangalam2023egoschema}. Recent benchmarks have pushed towards assessing reasoning for temporal questions~\cite{grunde2021agqa,xiao2021next,wu2021star_situated_reasoning}, longer videos~\cite{yu2019activitynet,mangalam2023egoschema}, and on domains like instructional~\cite{yang2021just} and egocentric videos~\cite{Gao_2021_ICCV,mangalam2023egoschema}. We evaluate our modular approach on four diverse and representative videoQA tasks: NExT-QA~\cite{xiao2021next}, iVQA~\cite{yang2021just}, EgoSchema~\cite{mangalam2023egoschema}, and ActivityNet-QA~\cite{yu2019activitynet}.

\para{End-to-end Models for VideoQA.} 
The recent success of LLMs~\cite{chowdhery2022palm,zhang2022opt,touvron2023llama,palm2} has led to an explosion of multimodal models that jointly understand vision and text data.
Many works map frozen image encoders~\cite{dosovitskiy2020image,clip2021,fang2023eva} to the LLM textual embedding space: \eg, Flamingo~\cite{alayrac2022flamingo}, via a Perceiver resampler~\cite{jaegle2021perceiver}, or BLIP2~\cite{li2023blip} and Video-LLaMa~\cite{damonlpsg2023videollama}, via Q-formers for audio/vision~\cite{fang2023eva,girdhar2023imagebind}. GIT2~\cite{wang2022git} and PALI~\cite{chen2022pali,chen2023pali,chen2023pali3} use simple encoder-decoder style architectures which are trained for image captioning, while MV-GPT~\cite{seo2022end} finetunes a native video backbone~\cite{arnab2021vivit} for video captioning.
Although trained with a generative (captioning) objective, such models achieve strong results for general vision-language tasks (cast as auto-regressive generation with question as prefix). More recent works such as Instruct-BLIP~\cite{dai2023instructblip}, MiniGPT-4~\cite{zhu2023minigpt}, and VideoBLIP~\cite{yu2023videoblip} improve zero-shot results with strong instruction tuning but generally, end-to-end methods can be difficult to interpret.

For videos in particular, memory limits in end-to-end models require significant downsampling: \eg temporally sampling a few frames with large strides~\cite{wang2022git, chen2023pali}, spatially subsampling each frame to a single token~\cite{yang2023vid2seq, zhou2018end, wang2021end}.
Such models also tend process each frame with equal importance. Unlike such works, our model has an explicit grounding stage which searches for the most relevant video frames to be processed in more detail. Other grounding works for videoQA include
SeViLa~\cite{yu2023self}, MIST~\cite{gao2023mist}, and NExT-GQA\cite{xiao2023can}; our model differentiates from these prior works by incorporating modular multi-stage reasoning.

\para{Visual Programming and Modularity.} Visual programming methods~\cite{subramanian2023modular,suris2023vipergpt,gupta2023visual,johnson2017inferring,andreas2016neural,cho2023visual} have shown promise towards addressing the limitations of end-to-end systems, by composing multiple sub-task specific modules into an executable program. Prior (earlier) work on neural modular networks~\cite{andreas2016neural,johnson2017inferring} made initial progress towards this goal, but were eventually outpaced by developments in large-scale end-to-end models. Recent work like CodeVQA~\cite{subramanian2023modular}, ViperGPT~\cite{suris2023vipergpt}, and VisProg~\cite{gupta2023visual} demonstrated accuracies on par with some end-to-end systems~\cite{li2023blip}, by replacing the event/language parsing with a code-finetuned LLM that can generate an entire python program (which invokes a number of provided APIs in the prompt). While these approaches are effective in terms of interpretability and flexibility in solving VQA, they share common limitations in that they heavily rely on a `single-prompt' with large, complex code generation examples~\cite{khandelwal2023analyzing}, which must generate the entire program without access to the image. 

\para{Multistage Planning Models.}
Recent methods have explored directly using natural language as the intermediate representation between large multimodal models. One emerging class of models are Socratic models~\cite{zeng2022socratic}, which use few-shot or zero-shot prompting of LLMs and VLMs to solve video tasks, \eg VidIL~\cite{wang2022language} which feeds image captions, frame attributes and ASR to an LLM to perform video-language tasks. The closest to our work is AVIS~\cite{hu2023avis}, which also uses multistage LLMs with an external memory for the task of visual question answering. However unlike AVIS which works on knowledge focused QA for images, we focus on the challenging domain of videos, where reasoning over multiple frames is required. A key difference therefore is our grounding stage, which determines which frames in a (potentially long) video contain the most relevant information to then deploy additional reasoning steps and tools in a more effective manner. 
\section{Technical Approach}
\label{sec:method}

In this section, we contextualize our technical approach for videoQA (Sec.~\ref{sec:method:task-vqa}) by discussing limitations in the standard single-stage (Sec.~\ref{sec:method:single-stage}) paradigm before presenting our main multistage modular reasoning model \ours{}(Sec.~\ref{sec:method:multi-stage}).

\subsection{Preliminaries: Video Question-Answering}
\label{sec:method:task-vqa}

\para{Task.} We focus on the task of video question-answering (VideoQA) as it provides a good testbed for video reasoning for multimodal systems.
Formally, we are given an input video $V = \{v_1, \dots, v_l \} $ with $l$ frames and a corresponding question $Q$ in natural language with a groundtruth answer $A$, either directly from the question alone \cite{yang2021just,yu2019activitynet}, or from among a set of candidate options $A \in A_{cands}$ \cite{xiao2021next,mangalam2023egoschema}.
The task is to develop a model $M$ such that:
\begin{equation}
	M(V, Q, [A_{cands}]) \mapsto A	
\end{equation}
\noindent where $A_{cands}$ are present for closed-set VQA settings \cite{mangalam2023egoschema,xiao2021next} and not present for open-ended VQA \cite{yang2021just,yu2019activitynet}.

\para{Design Approaches.} The approaches for addressing this task can vary broadly (Sec.~\ref{sec:related_work} and Fig.~\ref{fig:teaser});
here, we center our discussion around two key design principles in state-of-the-art systems for $M$:
(1) \textit{Modularity}, where individual, standalone modules focused on specific sub-tasks are leveraged, as opposed to a single monolithic black-box model; and
(2) \textit{Multi-stage planning}, where there are explicit intermediate outputs while the system determines \textit{which} modules to leverage and \textit{how} to use them most effectively, providing a more interpretable chain of execution. In this section, we focus on contrasting modular methods with single-stage (prior work) vs. multi-stage planning (our new model).

\subsection{Single-Stage Planning}
\label{sec:method:single-stage}

\para{Overview.} In Section~\ref{sec:related_work:visual_programming}, we discuss the broader space of visual programming and modular methods \cite{subramanian2023modular,johnson2017inferring,suris2023vipergpt}. Here, we focus on a specific representative state-of-the-art model (ViperGPT~\cite{suris2023vipergpt}) and discuss key limitations with its single-stage planning approach for modular videoQA, using notation consistent with prior work \cite{suris2023vipergpt,johnson2017inferring}.

\para{ViperGPT.} In the context of videoQA, ViperGPT is a system $M$ that consists of a single-stage program generator $\pi$ that takes as input the query $Q$ and a specialized prompt $P$ to directly output an intermediate executable program $z \in Z$, where $Z$ represents the space of all programs (Python, natural language, etc.). This program $z$ is then executed on the full input $(V,Q,[A_{cands}])$ to produce the final answer $A$. More formally, the full system can be described:
\begin{equation}
M_{\text{single-stage}}: \pi(Q, P) \mapsto z(V, Q, [A_{cands}]; L) \mapsto A
\end{equation}
\noindent where $L$ denotes the API module library used to construct the program $z$. The program generator $\pi$ is instantiated as a code-finetuned LLM \cite{palm2,chen2021evaluating} conditioned on a well-engineered prompt file $P$, consisting of two key components: (1) a custom API description with API usage examples, and (2) a set of dataset-specific program examples that illustrate how to translate the questions $Q$ found in the dataset distribution into a full program $z$ that composes these modules together effectively.

\para{Modules for Modular Reasoning.} ViperGPT and related models \cite{suris2023vipergpt,subramanian2023modular,gupta2023visual} leverage a specialized module library as described by their API to assemble executable programs $z$. We denote this library of API modules as $m \in L_{API}$: examples include open-vocabulary detection (OWL-ViT~\cite{minderer2022simple}), text-image scoring (CLIP~\cite{clip2021}, X-VLM~\cite{zeng2021xvlm}), and captioning (BLIP~\cite{li2023blip}). The overall program $z$ then describes the modular reasoning of the single-stage code generation LLM for a given query.

\para{Limitations.} While a single-stage approach suggests an appealing promise of simplicity, in practice, we observe that this design leads to brittle programs that do not produce reliable outputs\footnote{Also noted by concurrent analysis~\cite{khandelwal2023analyzing} in the image-language domain.}. We show a representative example for videos in Fig.~\ref{fig:qual_comparison_jcef}, with additional analysis in the supplement~\ref{sec_supp:additional_qual}.

The core issue of the overall system lies in the difficult task given to its single-stage planner: before performing visual reasoning, the model must output a full program without any grounding in the actual video itself. Thus, natural language ambiguity in the question cannot be resolved by visual context, important for video / event reasoning tasks~\cite{huang-buch-2018-finding-it,xiao2023can}.
Furthermore, by expecting the model to piece together full reasoning programs in one large LLM inference step, the necessary complexity of examples in the prompt grows accordingly. In practice, we observe this leads to the system overfitting on the specific examples provided (also noted by \cite{khandelwal2023analyzing}), falling short of realizing its true potential for compositional modular generalization. 

These limitations naturally beget \textit{two key questions}: \textbf{(Q1)} to what extent is brittle single-stage planning impacting accuracy, and \textbf{(Q2)} how well can we overcome these limitations through a \textit{multi-stage decomposition}? These motivate our proposed baseline and model in Sec.~\ref{sec:method:multi-stage}.

\begin{figure}[t]
    \begin{center}
        \includegraphics[width=0.99\linewidth]{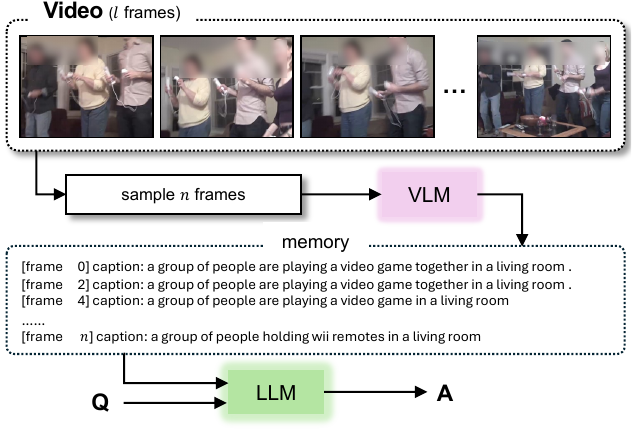}
    \end{center}
    \vspace{-5.0mm} 
      \caption{\textbf{A simple, strong baseline --} \oursjcef. Our proposed baseline consists of a zero-shot prompted vision-language model (VLM) which is used to caption $n$ uniformly sampled frames from a video ($n$ is all frames at 1FPS unless explicitly stated). These captions are then stored in an external memory, which is passed to a zero-shot prompted LLM that is used to answer a question about the video. We show that this baseline outperforms existing visual programming methods by a large margin and investigate ways to more effectively improve upon it in a modular, multistage manner.}
    \vspace{-6.0mm} 
\label{fig:jcef}
\end{figure}

\subsection{MoReVQA: Multi-stage, Modular Reasoning}
\label{sec:method:multi-stage}
\label{sec:method:simple-baseline}
\para{Motivation: A Simple, Strong Baseline (\oursjcef).}
To empirically characterize the limitations of single-stage approaches (per Q1), we create a simple but effective Socratic baseline called \jceffull{} (Figure~\ref{fig:jcef}), consisting of two strong modules $m_{VLM}$~\cite{chen2023pali3} and $m_{LLM}$~\cite{palm2}.
Our baseline involves no training, directly prompting these large off-the-shelf models in what can be considered as a very simple, deterministic ``program''.
For each video, we sample $n \leq l$ frames from the video $V$, captioning each frame with an image-language model $m_{VLM}$. These $n$ captions are then combined with frame numbers (e.g., ``[frame 5] caption: a person is throwing a baseball in a field'') into a prompt $P$ used to query the LLM $m_{LLM}$ with the question $Q$ and candidate answers $A_{cand}$ for multiple choice questions (prompt details in supplement~\ref{sec_supp:implementation_details}).
By comparing to a state-of-the-art baseline (ViperGPT+), upgraded with the \textit{same} modules $m_*$, we can observe the limitations of single-stage planning designs: surprisingly, \jcef{} \textit{outperforms} this single-stage baseline (Sec.~\ref{sec:experimental_results}).

\para{\ours{} Overview.}
We address our second key question (Q2) by considering a \textit{decomposition} of the single-stage pipeline into multiple stages, in order to effectively improve \textit{beyond} our \jcef{} baseline. Our new proposed model, multi-stage modular reasoning for videoQA (\ours), consists of three stages, rooted in key sub-tasks that are general to videoQA (and related video-language reasoning tasks) across benchmarks and domains: (1) \textit{event parsing} (understanding what is relevant in the input language), (2) \textit{grounding} (understanding what is relevant in the input video), and (3) \textit{reasoning} (understanding the relevant events, their attributes, and their relationships).

An overview of the pipeline is provided in Fig. \ref{fig:ours_modular_architecture}.
Each stage is distinct yet interconnected, employing an LLM that generates a set of API calls tailored for the specific subtasks.
Importantly, these APIs are backed by the same off-the-shelf pretrained models~\cite{minderer2022simple,clip2021,chen2023pali3} considered in the single-stage setting (Sec.~\ref{sec:method:single-stage}) for consistent comparison.
Central to this process is a shared external memory, managing and storing information across stages, including natural language events, grounded regions of the video, video captions, and intermediate tool outputs (details in  Sec. \ref{sec:implementation}).

Through this decomposition, our \ours{} model $M_{\text{multi-stage}} = \tightset{M_1, M_2, M_3}$ relies on smaller focused prompts $\tightset{P_1, P_2, P_3}$ for each stage\footnote{Please see supplement~\ref{sec_supp:implementation_details} for prompts and API details.}; furthermore, intermediate reasoning outputs $\tightset{z_1, z_2, z_3}$ are able to handle different aspects of the overall task, and incorporate grounding in the video itself to resolve ambiguities and inform new \textit{intermediate} reasoning steps in a more effective manner than the ungrounded single-stage setting. We describe each stage $M_i$ as follows:

\begin{figure*}
    \begin{center}
        \includegraphics[width=1.0\linewidth]{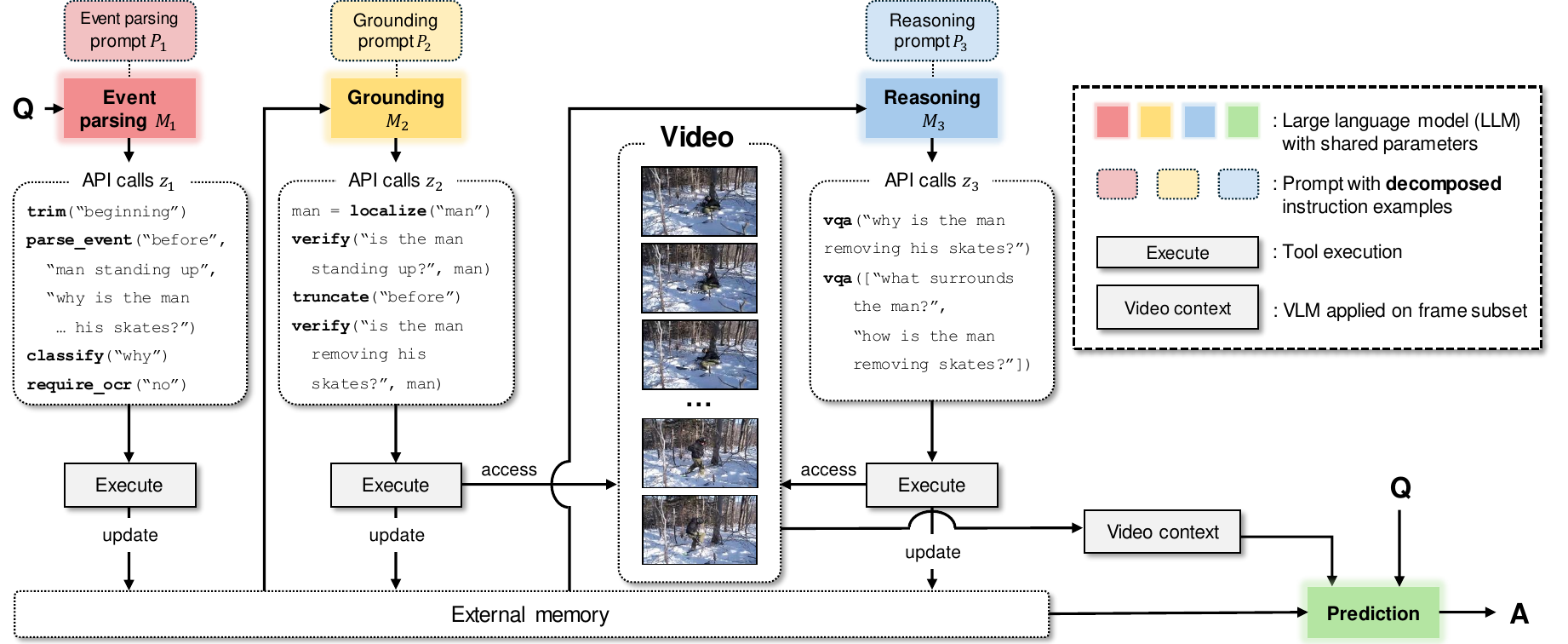}
    \end{center}
    \vspace{-5.0mm} 
\caption{\textbf{Modular Reasoning for Video Question-Answering (\ours).} To address the limitations of single-stage planning LLMs, we propose a new multi-stage, modular method $M_{\text{multi-stage}}$ that decomposes planning and execution into three key steps, motivated by sub-tasks inherent to videoQA: (i) event parsing $M_1$, (ii) grounding $M_2$, and (iii) reasoning $M_3$. See Section~\ref{sec:method:multi-stage} for additional details.
      }
    \vspace{-2.0mm} 
\label{fig:ours_modular_architecture}
\end{figure*}

\noindent\textbf{Event parsing stage $M_1$.}
The first stage focuses on the initial analysis and processing of the input question $Q$. Different from traditional language parsing in early modular systems~\cite{johnson2017inferring,andreas2016neural}, our $M_1$ stage parses at the \textit{event}-level rather than word-level, focusing on higher-level video semantics while still decomposing relationships and attributes for later stages. 
Our event parsing prompt $P_1$ (see supplementary~\ref{sec_supp:implementation_details}) conditions the LLM to examine the input question, perform parsing tasks such as detecting temporal hints and relationships (``in the beginning of the video'', ``before'', ``during''), sub-question types (location, description, explanation), and whether the language would suggest additional tool types (\eg OCR).
The LLM then produces a set of API calls based on these parsing tasks, expressed as $z_1$, which when executed populates the external memory with relevant language-only data for later stages.

\noindent\textbf{Grounding stage $M_2$.}
In this stage, the focus shifts to \textit{grounding} identified events, a critical process to help resolve ambiguities and direct tool-use in the final reasoning stage to the temporal regions where they can be most effective.
Here, the prompt $P_2$ is constructed with the external memory state with outputs from $M_1$ (\eg, parsed events), and conditions the LLM to identify candidate frames and temporal regions in the video with vision-language modules $m$ for entity detection and image-text alignment. The resulting $z_2$ is then executed on the video, and the output grounding (spatially and temporally) is appended to the external memory. Importantly, this process includes API calls designed to help verify and resolve event ambiguity through visual grounding, as illustrated in Fig.~\ref{fig:ours_modular_architecture}.

\noindent\textbf{Reasoning stage $M_3$.}
The final stage of our system performs grounded reasoning before the final prediction. The LLM prompt $P_3$ is based on the memory state after the previous two stages, and constructs a final $z_3$ executable with API calls (Fig.~\ref{fig:ours_modular_architecture}) designed around reasoning sub-questions to unravel different aspects of the original question, and focusing vision-language modules on the specific grounded regions of the video identified previously.
This localized, context-specific information is subsequently combined with a more general $n \leq l$ captions from frames sampled uniformly (\textit{general video context}, in Fig.~\ref{fig:ours_modular_architecture}) across the video to form a comprehensive (temporally-sorted) basis for a final \textit{prediction} LLM to output the final answer $A$ (in general, $n$ here is significantly less than with JCEF). This final API call here corresponds to the standard \texttt{llm\_query} module found in prior work \cite{suris2023vipergpt}, typically at the end of the program to ensure correct formatting and candidate answer selection.

\noindent\textbf{Flexibility and Memory.}
The modular architecture of \ours{} allows it to be dynamically tailored to a wide range of datasets, question types, and tasks by selectively engaging different APIs and reasoning strategies based on the task at hand.
In particular, simple questions beget a ``simpler'' execution pipeline (stages are equipped with ``no-op'' like behavior, if necessary), while more complex questions are processed with a complex instruction set.
This adaptability is facilitated by the external memory component, which not only serves as a repository of information across stages but also enables the system to iteratively refine its understanding and approach based on the evolving context.
We highlight that each stage (planning and execution) are informed by previous stages through this memory, which leads to more robust reasoning behavior.

\section{Experiments}
\label{sec:experimental_results}
Here, we describe the VideoQA datasets and evaluation metrics used (Sec. \ref{sec:datasets}), our key baselines (Sec. \ref{sec:baselines}) and implementation details (Sec. \ref{sec:implementation}), and our discussion of results and analysis (Sec. \ref{sec:results}).

\subsection{Datasets and Evaluation Metrics} \label{sec:datasets}
We consider four standard videoQA benchmarks to assess the efficacy of our proposed method, across a range of representative video domains, lengths, and question types.

\para{NExT-QA~\cite{xiao2021next}} is focused on understanding the ability of videoQA systems to effectively answer questions across three types: causal (C), temporal~(T) and descriptive~(D). We focus on the same multiple choice (MC) setting reported in prior single-stage modular reasoning work \cite{suris2023vipergpt}, where each video clip (avg length, 43s) contains one question and 5 candidate answers; we use 4996 val video-question pairs.

\para{iVQA~\cite{yang2021just}} consists of 7-30s instructional video clips sampled from the HowTo100M dataset~\cite{miech19howto100m}, with 5615 training and 
1879 testing clips (after removing clips no longer online). Each clip has a question and annotated set of ground truth answers. We note that iVQA is open-ended (OE) videoQA, and no candidate answers are provided as input.

\para{EgoSchema~\cite{mangalam2023egoschema}} is a recent dataset of \textit{long} egocentric videos (180s) based on the Ego4D~\cite{grauman2022ego4d} benchmark with multiple-choice (MC) questions, designed specifically to assess long video understanding. EgoSchema is focused entirely on evaluation: the hidden test set consists of 5000 videos via an evaluation server, of which 500 were publicly released for validation. We report results (accuracy) on the main (full, 5k) test set for comparison with prior work.

\para{ActivityNet-QA~\cite{yu2019activitynet}} has 5800 videos, each accompanied by 10 annotated question-answer pairs to characterize model comprehension of actions, objects, locations, and events. ActivityNet-QA is open-ended (like iVQA) with long videos (180s avg., like EgoSchema). We report test set results using GPT-based evaluation following~\cite{maaz2023videochatgpt,li2024videochat,zhang2023llamaadapter}.

\subsection{Baselines} \label{sec:baselines}
We compare our method against a key set of baselines:

\para{Single-stage Planning (ViperGPT+).} As a representative state-of-the-art baseline for single-stage planning and modular reasoning, we reimplement ViperGPT\cite{suris2023vipergpt}, as described in Sec.~\ref{sec:method:single-stage}. We extend the open-source implementation and upgrade some of the modules/APIs to ensure consistent comparisons with our method and to replace prior module components that are not available (eg. GPT-3 Codex~\cite{chen2021evaluating}); full description in the supplement~\ref{sec_supp:implementation_details}. We evaluate this baseline on video datasets that were not used in the original paper (iVQA, EgoSchema, ActivityNet-QA) to better characterize single-stage planning for videoQA. 

\para{\fulljcef.} We also consider our \oursjcef{} baseline described in Sec.~\ref{sec:method:simple-baseline} as a simple but powerful Socratic model that is a step up in interpretability to a purely end-to-end system, but lacks the kind of modular compositionality that is present in more fully fledged modular reasoning systems. The VLM and LLM models used here are the same as with ViperGPT+ and our full system, for consistent comparison (details in Sec. \ref{sec:implementation}).

\para{Language-only baseline.}
We also compare our model with a language-only baseline, which is an LLM \cite{palm2} prompted to answer questions without any visual inputs, as a way to quantify the amount of non-visual language and/or common sense bias in each dataset. For consistent comparison, this language model is used across all modular methods.

\begin{table}[t]
    \begin{center}
    \scalebox{0.75}{
         \begin{tabular}{ccccc}
                \toprule
                 
                 \multirow{2}{*}{Method} & \multicolumn{4}{c}{Accuracy (\%)} \\\cline{2-5}\\[-2ex]
                  & NExT-QA & iVQA & EgoSchema & ActivityNet-QA\\

                 \midrule
                 Random (for MC) & 20.0 & - & 20.0 & - \\
                 LLM-only~\cite{palm2} & 48.5 & 15.0 & 41.0 & -\\
                 ViperGPT~\cite{suris2023vipergpt}
                 & 60.0 & - & - & - \\
                 \midrule
                 ViperGPT+ & 64.0 & 46.6 & 49.3 &  37.1 \\
                 \oursjcef         & \underline{66.7} & \underline{56.9} & \underline{49.9} & \underline{43.3} \\
                 \ours            & \textbf{69.2} & \textbf{60.9} & \textbf{51.7} & \textbf{45.3}\\
                 
                 \bottomrule
        \end{tabular}
    }
    \vspace{-2.0mm}
    \caption{\label{table:ablation_program_generation}
    \textbf{Comparison to single-stage modular methods.}
    ViperGPT~\cite{suris2023vipergpt} represents the state-of-the-art single-stage modular question answering system, and ViperGPT+ is our upgraded reimplementation for consistent comparison. Our \jcef{} strong performance highlights the relative weakness in single-stage planning models, which can lead to brittle programs and outputs. We find that our \ours{} model outperforms all key baselines.
    }
    \vspace{-5.0mm}
    \end{center}
\end{table}

\begin{figure*}[t]
    \begin{center}
        \includegraphics[width=0.99\linewidth]{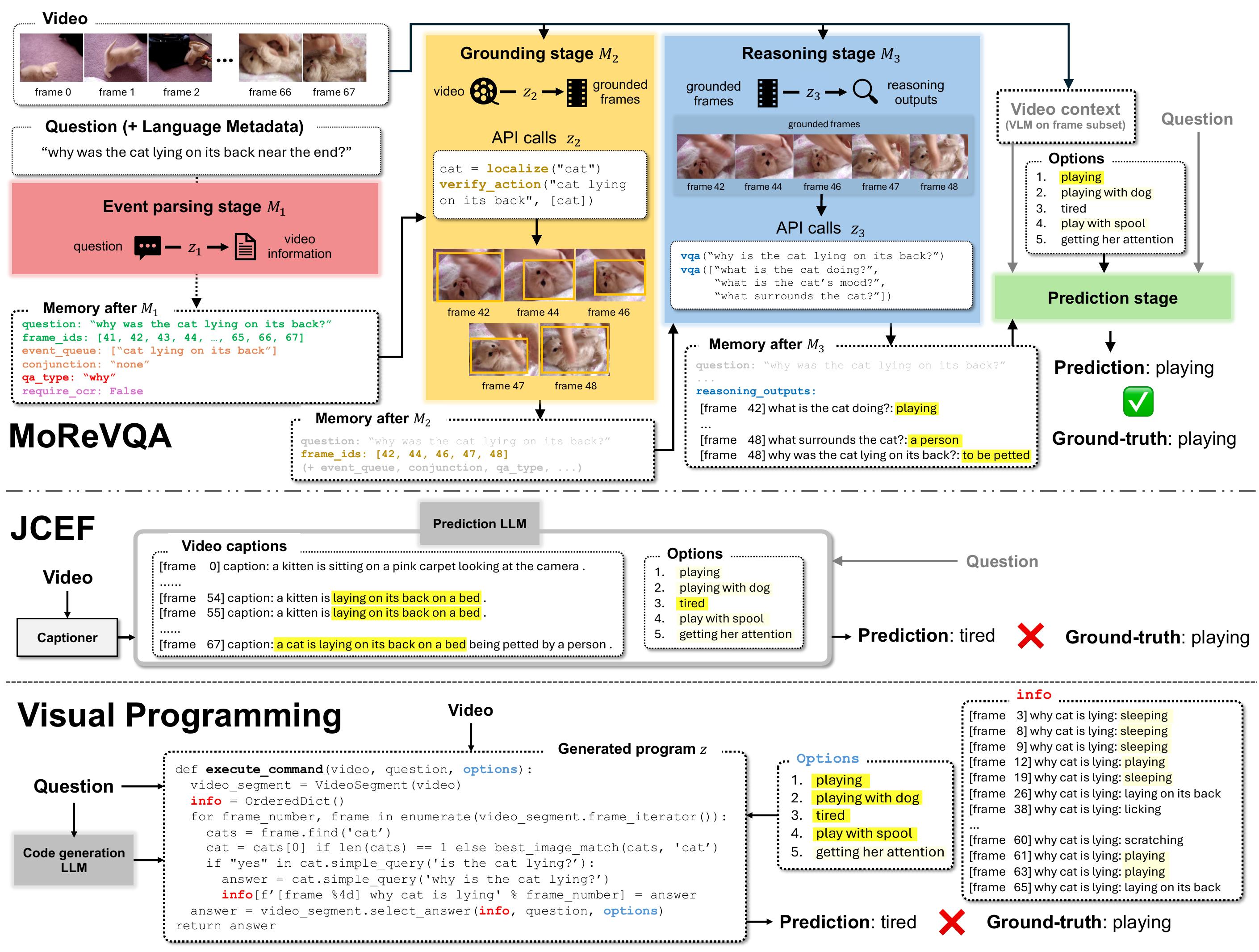}
    \end{center}
    \vspace{-5.0mm} 
\caption{\textbf{Example qualitative result of \ours{}} on NExT-QA. We observe that the intermediate outputs from our \ours{} model are interpretable: event parsing stage parses key events from language, and other tool-use metadata. The grounding stage then determines which frames contain the `cat lying on its back', and the reasoning stage reasons about relevant sub-questions for the final answer, which when combined with general video-level context (subset of frame captions), gives us the final correct answer. We observe that JCEF and ViperGPT+ fail to predict correct answer for the same example (Sec.~\ref{sec:results:analysis}); we provide more examples and analysis in the supplement~\ref{sec_supp:additional_qual}.
}
    \vspace{-3.0mm}
    \label{fig:qual_comparison_jcef}
\end{figure*}

\subsection{Implementation Details}
\label{sec:implementation}

Across all of our baselines and proposed models (\eg, \ours, \jcef, ViperGPT+), our core VLM is PALI-3 (5B)~\cite{chen2022pali} for image captioning and related APIs, and our core LLM is PaLM-2\cite{palm2} (\eg, every LLM stage in \ours{}, \jcef{}, the language-only baseline, and for the \texttt{llm\_query} module in ViperGPT+), unless otherwise specified.
Our video context / captioner component for MoReVQA considers $n=16$ uniformly sampled video frames as a default.
For \jcef{}, $n = l$, the number of frames in the video (at 1 frame per second); we provide additional \jcef{} ablations for different values of $n$ in the supplement~\ref{sec_supp:additional_results}.
We set decoding temperature to $0$ to match prior work \cite{suris2023vipergpt}; other base models and settings (\eg, OWL-VIT~\cite{minderer2022simple}, CLIP~\cite{clip2021}, etc.) for \ours{} and ViperGPT+ are in the supplement~\ref{sec_supp:implementation_details} and are also consistent wherever applicable. Our implementation relies on JAX/Scenic \cite{jax2018github,dehghani2021scenic}.

\ours{} stores outputs of each stages in an external memory system, backed by global variables for tracking and updating information through the model's processing stages.
These stages execute different API calls, \eg, event parsing reduces frame data for efficiency, the grounding stage focuses on object localization and action verification, and the reasoning stage decomposes and addresses the question with VQA on selected frames.
Additional API, memory, and LLM prompt details for \ours{} and other models are provided in supplementary~\ref{sec_supp:implementation_details}.

\begin{table}[t]
    \begin{center}
    \scalebox{0.85}{
        \begin{tabular}{ccccc}
                \toprule
                 
                 \multicolumn{3}{c}{Stages} & NExT-QA & iVQA \\\cline{1-3}\\[-2ex]
                 Event parsing & Grounding & Reasoning & Val & Test \\
                 \midrule
                 \xmark & \xmark & \xmark & 66.7 & 56.9 \\
                 \cmark & \xmark & \cmark & 68.3 & 56.9 \\
                 \cmark & \cmark & \xmark & 68.7 & 57.5 \\
                 \cmark & \cmark & \cmark & \textbf{69.2} & \textbf{60.9} \\
                 \bottomrule
        \end{tabular}
    }
    \vspace{-1.2mm}
    \caption{\label{table:ablation_multistage}\textbf{Ablation study of the various stages in~\ours.} We show the impact of each of the key stages of our proposed design, highlighting the improvements between a system without our proposed stages (top row; defaults to the JCEF baseline) and our multi-stage reasoning setting (bottom row; all 3 stages). We observe stages provide complementary (\eg, NExT-QA) and synergistic (\eg, iVQA) gains (additional ablations in supplement~\ref{sec_supp:additional_results}).
    }
    \vspace{-7.7mm}
    \end{center}
\end{table}

\subsection{Results and Discussion}\label{sec:results}

\subsubsection{Comparison to Baselines and Analysis}\label{sec:results:analysis}
We compare our method to the baselines outlined in Sec.~\ref{sec:baselines} in Table \ref{table:ablation_program_generation}.
The LLM-only baseline performance serves as a measure of language and commonsense bias: our results for NExT-QA and iVQA align with prior expectations (\eg, iVQA was explicitly designed to mitigate language bias).
However, this baseline also does surprisingly well on EgoSchema, in spite of its explicit emphasis on testing long form video understanding. We believe this could potentially be an artifact of its dataset construction process, which leveraged automatic LLM generations to form the question/candidate answer language inputs~\cite{mangalam2023egoschema} and may have introduced unintended language bias.

\begin{table*}
\subcaptionbox{NExT-QA~\cite{xiao2021next}}{
    \scalebox{0.71}{
    \begin{tabular}{ccc}
                \toprule
                 
                 Method & Val  & FT \\

                 \midrule
                 
                 \deemph{MIST-CLIP~\cite{gao2023mist}}
                 & \deemph{57.2} &  \multirow{3}{*}{\cmark} \\

                 \deemph{HiTeA~\cite{ye2023hitea}}                            
                 & \deemph{\underline{63.1}}  &   \\
                 
                 \deemph{SeViLa~\cite{yu2023self}}
                 & \deemph{\textbf{73.8}} &  \\

                 \midrule
                 \midrule
                 
                 ViperGPT~\cite{suris2023vipergpt}                
                 & 60.0 & \multirow{6}{*}{\xmark} \\
                 
                 BLIP-2$^{\text{concat}}$~\cite{li2023blip}
                 & 62.4 & \\
                 
                 BLIP-2$^{\text{voting}}$~\cite{li2023blip}
                 & 62.7 &  \\
                 SeViLA~\cite{yu2023self}
                 & 63.6 &  \\\cline{1-2}\\[-2ex]
                 \oursjcef                      & \underline{66.7}  & \\
                 \ours                          & \textbf{69.2} & \\
                 \bottomrule
    \end{tabular}}
}
\hfill
\subcaptionbox{iVQA~\cite{yang2021just}}{
    \scalebox{0.71}{
        \begin{tabular}{cccc}
                \toprule

                Method & Test & FT \\
                \midrule

                \deemph{VideoCoCa~\cite{yan2022videococa}} & \deemph{39.0}  & \multirow{3}{*}{\cmark} \\
                \deemph{FrozenBiLM~\cite{yang2022zero}} & \deemph{\underline{39.7}}  &  \\
                \deemph{Text+Text}~\cite{lin2023towards} & \deemph{\textbf{40.2}}  &  \\

                \midrule
                \midrule
                FrozenBiLM~\cite{yang2022zero} & 27.3 & \multirow{6}{*}{\xmark} \\
                BLIP-2$_\text{\small{(FlanT5XXL)}}$~\cite{li2023blip}  & 45.8 &  \\
                InstructBLIP$_\text{\small{(FlanT5XL)}}$~\cite{dai2023instructblip}  & 53.1 &  \\
                InstructBLIP$_\text{\small{(FlanT5XXL)}}$~\cite{dai2023instructblip} & 53.8 &  \\\cline{1-2}\\[-2ex]
                \oursjcef  & \underline{56.9} &  \\ 
                \ours  & \textbf{60.9} &  \\
                \bottomrule
        \end{tabular}}
}
\hfill
\subcaptionbox{EgoSchema~\cite{mangalam2023egoschema}}{
    \scalebox{0.76}{
        \begin{tabular}{ccc}
                \toprule

                Method & Test & FT \\
                \midrule
                VIOLET~\cite{fu2021violet} & 19.9 & \multirow{9}{*}{\xmark} \\
                SeViLA~\cite{yu2023self} & 22.7 & \\
                FrozenBiLM~\cite{yang2022zero}       & 26.9 &  \\
                mPLUG-Owl~\cite{ye2023mplug}        & 31.1 & \\
                InternVideo~\cite{wang2022internvideo}      & 32.1 & \\
                $^{*}$ShortViViT~\cite{papalampidi2023simple} & 31.0 & \\
                $^{*}$LongViViT~\cite{papalampidi2023simple}      & 33.3 & \\
  
                \cline{1-2}\\[-2ex]
                \oursjcef & \underline{50.0} &  \\ 
                \ours   & \textbf{51.7} & \\
                \bottomrule
        \end{tabular}}
}
\hfill
\subcaptionbox{ActivityNet-QA~\cite{yu2019activitynet}}{
    \scalebox{0.73}{
        \begin{tabular}{cccc}
                \toprule

                Method & Test & FT \\
                \midrule

                Video-LLaMA~\cite{damonlpsg2023videollama} & 12.4 & \\
                VideoChat~\cite{li2024videochat} & 26.5 & \multirow{6}{*}{\xmark}  \\
                $^{*}$LLaMa adapter~\cite{zhang2023llamaadapter} & 34.2 &  \\ 
                $^{*}$Video-ChatGPT~\cite{maaz2023videochatgpt} & 35.2 &  \\  
                ViperGPT+ & 37.1 &  \\
                \cline{1-2}\\[-2ex]
                \oursjcef & \underline{43.3} &  \\
                \ours & \textbf{45.3} &  \\

                \bottomrule
                \vspace{3mm}
        \end{tabular}
    }
}
\hfill
\vspace{-2.0mm}
\caption{\label{table:sota_comparison}\textbf{Comparison to SOTA on the standard video question-answering datasets:} (a) NExT-QA, (b) iVQA, (c) EgoSchema, and (d) ActivityNet-QA. Our method \ours~outperforms all training-free prior work or exceeds prior state-of-the-art fine-tuned systems (in grey), on the main validation datasets~\cite{xiao2021next,yang2021just,mangalam2023egoschema,yu2019activitynet}. FT indicates fine-tuned methods. Methods with asterisk $^{*}$ indicate concurrent work.
}
\vspace{-5.0mm}
\end{table*}

Next, we highlight our simple Socratic baseline \oursjcef{} outperforms state-of-the-art single-stage programming methods such as ViperGPT+ (our upgraded implementation) across all datasets, even though both baselines have access to the \textit{same} VLM and LLM modules. This gives us a quantitative assessment of the impact of brittle program generations and tool executions for the single-stage model, as observed in our qualitative analysis (Figure~\ref{fig:qual_comparison_jcef}, bottom + additional examples in supplement~\ref{sec_supp:additional_qual}); we also highlight concurrent analysis \cite{khandelwal2023analyzing} which found similar failure modes in image-language settings.

Finally, we note that our model \ours{} outperforms all previous training-free baselines across all datasets, while at the same time providing intermediate interpretable outputs -- in Fig. \ref{fig:qual_comparison_jcef}, we show a representative example on NExT-QA. We also show limitations in both JCEF and ViperGPT given the same example: while JCEF is a strong baseline, the general-level captions are not always informative enough to answer the question, while the program generated by ViperGPT+ does not focus on the frames in the correct part of the video (specifically, the ``if'' statement condition erroneously triggers on irrelevant early frames, resulting in misleading ``info''). See supplement~\ref{sec_supp:additional_qual} for more.

We also ablate the stages in~\ours{} (Table \ref{table:ablation_multistage}).
Our ablation without any stages (top row) effectively defaults to the JCEF baseline with only frame-level captions and a final LLM prediction stage.
The absence of grounding indicates that we simply return a single middle frame for this stage.
For all ablations without the final reasoning stage, we retain a final LLM prediction on top of the shared memory state, \ie, the reasoning performs the final VQA only with the input question (without supporting questions).
We observe all three stages of our model (event parsing, grounding, and reasoning) provide complementary (\eg on NExT-QA) and synergistic (\eg on iVQA) gains, and meaningfully improve over the JCEF baseline.
The added benefit is the interpretability of the intermediate outputs stored in the external memory.
Further, we ablate the impact of key components in the memory; when the original question is provided to the reasoning stage, as opposed to a revised version (\eg, \texttt{question} in Fig.~\ref{fig:qual_comparison_jcef}), we note accuracy drops of 1.3\% and 3.9\% on NExT-QA and iVQA respectively.
When grounded frame locations are only given to prediction stage, instead of reasoning stage, we observe drops of 1.2\% and 3.9\% on the same datasets. Additional examples and analysis are in the supplement~\ref{sec_supp:additional_results}, including specific API usage statistics.

\vspace{-3.0mm}
\subsubsection{Comparison to State-of-the-Art}
Finally, we compare our method to the state-of-the-art methods on four datasets -- NExT-QA, iVQA, EgoSchema, and ActivityNet-QA (Table~\ref{table:sota_comparison}) -- in which the numbers in bold and underline respectively indicate the best and second best.
On all datasets we outperform previous zero-shot and few-shot methods by large margins -- on NExT-QA we outperform SeViLA by almost 6\%, making progress towards closing the gap to fully finetuned performance. On iVQA we outperform the nearest method InstructBLIP~\cite{dai2023instructblip} by almost 7\%, while on EgoSchema the gaps are the largest (approx. 20\%). For EgoSchema, we report results using $n=30$ video frames; we provide results with other values of $n$ in the supplement~\ref{sec_supp:additional_results}. We also show strong results on ActivityNet-QA, outperforming concurrent work \cite{zhang2023llamaadapter,maaz2023videochatgpt} under consistent evaluation protocols.

\para{Extensions to related tasks.} In our supplement~\ref{sec_supp:additional_results}, we describe extensions of our \ours{} system to other tasks. We consider grounded videoQA (localizing the relevant video segment while providing the answer) on the recent NExT-GQA~\cite{xiao2023can} dataset, and highlight our \textit{training-free} \ours{} achieves strong performance (44.1 mIoP / 31.0 Acc@GQA) vs. prior state-of-the-art SeViLa (29.5 mIoP / 16.6 Acc@GQA; trained with grounding annotations). We also show on ActivityNet-Para \cite{krishna2017dense} strong performance for video paragraph captioning (\ours{} 28.2 CIDEr vs. finetuned SOTA Vid2Seq \cite{yang2023vid2seq} with 28.0), even though our method is \textit{training-free}. We observe our system's reasoning enables diverse long captions of human-centric events.
\section{Conclusion}
\label{sec:conclusion}

In this work, we have presented a baseline (JCEF) to help characterize limitations with single-stage planning models, along with \ours{}, a new, modular, and decomposed multi-stage pipeline for video question answering.
Our framework consists of $3$ stages -- event parsing, grounding, and reasoning with an external memory. \ours{} achieves state-of-the-art results on popular VideoQA benchmarks, while producing interpretable intermediate outputs.
Refer to supplement~\ref{sec_supp:conclusion} for limitations and broader impacts.

\para{Acknowledgements.}
We sincerely thank ViperGPT~\cite{suris2023vipergpt} authors for sharing additional details helpful for the development of ViperGPT+, and grateful to Chen S., Jasper U., and Lluis C. for discussions.
Minsu Cho acknowledges IITP grant (2022-0-00959: ``Few-shot learning of causal inference in vision and language'') support by Korea (MSIT).

{
    \small
    \bibliographystyle{ieeenat_fullname}
    \bibliography{egbib}
}

\maketitlesupplementary

In this supplementary material, we provide additional implementation details with the prompts used in the experiments (Sec.~\ref{sec_supp:implementation_details}), more results and analyses (Sec.~\ref{sec_supp:additional_results}), and additional qualitative comparisons (Sec.~\ref{sec_supp:additional_qual}).
Finally, we conclude this supplement by discussing broader impacts, limitations, and future work in Sec.~\ref{sec_supp:conclusion}. Additional visualizations and expanded supplementary material can be found in the project website\footnote{Project website: {\tt\url{http://juhongm999.github.io/morevqa}}}.

\appendix

\newcommand{\beginsupplement}{
        \setcounter{table}{0}
        \renewcommand{\thetable}{A\arabic{table}}
        \setcounter{figure}{0}
        \renewcommand{\thefigure}{A\arabic{figure}}
     }
\beginsupplement

\lstset{
  basicstyle=\fontsize{7pt}{8pt}\selectfont\ttfamily,
  breaklines=true,
  tabsize=4,
  showstringspaces=false,
  numbers=none,
  xleftmargin=2em,
  captionpos=b,
  commentstyle=\color{black},
  keywordstyle=\color{black},
  stringstyle=\color{black},
  extendedchars=true,
  escapeinside={(*@}{@*)}
}
\renewcommand{\lstlistingname}{Prompt}
\renewcommand{\thesection}{\Alph{section}}

\section{Implementation details}
\label{sec_supp:implementation_details}
In this section, we present additional implementation details on our method MoReVQA.
Specifically, we focus on the API functions used for the grounding and reasoning stages.
In grounding stage, two functions are used for spatio-temporal localization: \texttt{localize()} and \texttt{verify\_action()};
\texttt{localize()} function takes natual lanuage objects as input, utilizing two pretrained image-text models:
OWL-ViT~\cite{minderer2022simple} which employs ViT-Base CLIP model with 32-patch size and a maximum query length of 16 tokens as an image-text embedding network and CLIP~\cite{clip2021} which uses ResNet-50 backbone~\cite{he2015resnet} with text-image comparison threshold of 0.7.
\texttt{verify\_action()} function employs a vision-language model PALI-3 (5B)~\cite{chen2022pali}, designed to validate the action of a single input object or interactions among multiple objects.
PALI-3 is also integral to the reasoning stage, where it facilitates VQA on grounded frames.
Note that all the three vision-language models, OWL-ViT, CLIP, and PALI-3, utilized in our MoReVQA framework are pretrained in the image domain, not the video domain, suggesting a potential avenue for further improvements.
Furthermore, we re-iterate that all base models used across all methods (ours + key baselines) are consistent, as discussed in the main paper (\eg, ViperGPT+ also uses the same OWL-ViT, PALI-3, etc. where applicable).

Outputs from different stages of ~\ours{} are stored and retrieved via an external memory system. We highlight $6$ variables, commonly used across input examples, here:
 \texttt{frame\_ids} (adaptively updated as objects within the video are localized during grounding), 
 \texttt{question} (tracking transformations of the core question $Q$ during parsing, grounding, and reasoning), 
 \texttt{event\_queue} and \texttt{conjunction} (extracted in event parsing and utilized in grounding),
 \texttt{qa\_type} (for tracking question sub-type, extracted in event parsing and utilized during reasoning), and \texttt{require\_ocr} (adjusts the prefix in the core PALI-3 call, to encourage OCR outputs if necessary; set in event parsing). As seen in Figures~\ref{fig:supp_qual_4703} - \ref{fig:supp_qual_e350}, at each stage, different API calls are executed by ~\ours~: 
 event parsing involves video trimming (\eg truncating `\texttt{frame\_ids}' by 40\% for parsed temporal arguments `beginning', `middle' or `end' for efficiency), event parsing, question classification, and OCR checks;
 the grounding stage focuses on localizing objects and verifying their actions;
 and the reasoning stage decomposes questions into related questions and then invokes VQA on the specifically identified frames, where the line of reasoning is more relevant. 

We also note that, since the original submission of this work for review, there have been significant developments in stronger VLM and LLM modules and components \cite{yang2023dawn,openai2024gpt4,geminiteam2023gemini} -- while our focus is on studying (in a controlled manner, with consistent baseline) the design space for modular reasoning models, inclusion of such models would likely further increase our ``state-of-the-art'' numbers for the benchmarks examined; we leave this to future work.

\noindent\textbf{Prompts.}
On our project website, we provide the prompts with partial example sets used in the proposed method MoReVQA and Visual Programming method, \eg, ViperGPT+. We describe them further here:
Prompts 1, 2, and 3 are used for event parsing, grounding, and reasoning stages of MoReVQA respectively. Short programs and plans generated as part of each stage are executed and interact with the shared memory state across all three phases. Importantly, our system has the capability to dynamically decide to what extent each stage is useful to the question (and can implicitly skip stages when necessary; see Sec.~\ref{sec_supp:additional_qual} for examples of such ``no-op'' operations).
After the three stages, MoReVQA predicts the final answer by utilizing Prompts 4 and 5 for multiple-choice (NExT-QA and EgoSchema) and open-ended (iVQA and ActivityNet-QA) questions respectively.

Based on the ViperGPT~\cite{suris2023vipergpt} original prompt, which the authors provided directly (including their examples), we provide the prompt that we improved for the evaluation of ViperGPT+ (which also has the upgraded, consistent base module set used in JCEF and \ours).
The examples of the ViperGPT+ prompt are similar to its predecessor while integrating improvements to ensure precise entity grounding with additional if/else statements.
Across all systems, models are backed by Python implementation/execution, and underlying base models/implementations are consistent across all key methods and baselines that we run ourselves, as described in the main paper.
\section{Additional results and analyses}
\label{sec_supp:additional_results}

In this section, we present additional results ans analyses to further support the findings reported in the main paper.
These include expanded comparisons on the NExT-QA and EgoSchema datasets, an ablation study on JCEF and further evaluation of \ours{} on other tasks.

\noindent\textbf{Expanded NExT-QA SOTA comparison.}
In Tables~\ref{table:nextqa_sota_comparison_ex1} and~\ref{table:nextqa_sota_comparison_ex2}, we expand SOTA comparison table on NExT-QA dataset initially presented in the main paper.
Table~\ref{table:nextqa_sota_comparison_ex1} shows a expanded comparison with a particular focus on the Temporal and Causal subsets to highlight how each model performs in specific types of videoQA;
MoReVQA shows its superiority over all the baselines, outperforming previous state of the art by approximately 3\% and 9\% in respective Temporal and Causal subsets.
Table~\ref{table:nextqa_sota_comparison_ex2} focuses on the more challenging subsets of Causal and Temporal subsets, \eg, Hard-C and Hard-T.
our proposed method MoReVQA again outperforms its counterparts, achieving a new state of the art in Hard-C (63.42\%) and Hard-T (59.57\%), showcasing advanced grounding and reasoning of MoReVQA.

\noindent\textbf{Expanded EgoSchema SOTA comparison.}
For a comparative analysis of how the model performs under varying degrees of available video information, we evaluate \ours{} on EgoSchema dataset under different numbers of uniformly sampled frames, specifically 30 and 90.
In 30-frame setting, \ours{} achieves 51.7\%, setting a new state of the art with substantial improvements over the previous baseline~\cite{wang2022internvideo} which stood at 31.8\% under identical conditions.
A similar trend was observed in the 90-frame scenario, with \ours{} recording 51.1\%, once again surpassing the baseline~\cite{wang2022internvideo} which noted 32.1\% in the same setting.
The results collectively indicate the superiority of \ours{} across varied frame-sample settings.

Moreover, the marginal performance difference between the 30- and 90-frame settings (51.7\% {\em vs.} 51.1\%) suggests a plateau in the use of additional video information in this dataset.
The incremental frame count from 30 to 90 frames did not result in a marked improvement, the same finding that aligns with observations from the original EgoSchema study by Mangalam \etal~\cite{mangalam2023egoschema}.
This outcome may indicate that the video data within this dataset contain substantial redundancy, to the extent that a set of 30 (or fewer) uniformly sampled frames can be sufficient to accurately predict the correct answers.

\begin{table}[t]
    \begin{center}
    
    \scalebox{0.85}{
        \begin{tabular}{ccccc}
                \toprule

                \multirow{2}{*}{Method} & \multicolumn{4}{c}{NExT-QA accuracy (\%)}  \\
                & Average & Temporal & Causal & Fine-tuned \\
                \midrule
                \deemph{All-in-One~\cite{wang2022one}}
                & \deemph{50.6} & \deemph{48.6} & \deemph{48.0} & \multirow{9}{*}{\cmark} \\
                 
                \deemph{MIST-CLIP~\cite{gao2023mist}} 
                & \deemph{57.1} & \deemph{56.6} & \deemph{54.6} & \\
                
                \deemph{HiTeA~\cite{ye2023hitea}}
                & \deemph{63.1} & \deemph{58.3} & \deemph{62.4} & \\

                \deemph{VGT~\cite{xiao2022video}} & \deemph{55.0} & \deemph{55.1} & \deemph{52.3} & \\
                
                \deemph{VFC~\cite{momeni2023verbs}} & \deemph{58.6} & \deemph{53.3} & \deemph{57.6} & \\
                
                \deemph{InternVideo~\cite{wang2022internvideo}}
                & \deemph{63.2} & \deemph{58.5} & \deemph{62.5} & \\
                 
                \deemph{BLIP-2$^{\text{voting}}$~\cite{li2023blip}}
                & \deemph{70.1} & \deemph{65.2} & \deemph{70.1} & \\

                \deemph{BLIP-2$^{\text{concat}}$~\cite{li2023blip}}
                & \deemph{\underline{72.6}} & \deemph{\underline{68.1}} & \deemph{\underline{72.9}} & \\
                 
                \deemph{SeViLa~\cite{yu2023self}} 
                & \deemph{\textbf{73.8}} & \deemph{\textbf{69.4}} & \deemph{\textbf{74.2}} & \\
                 
                \midrule
                \midrule
                InternVideo~\cite{wang2022internvideo}
                & 49.1 & 43.4 & 48.0 & \multirow{8}{*}{\xmark} \\
                
                VFC~\cite{momeni2023verbs}
                & 51.5 & 45.4 & 51.6 & \\

                BLIP-2$^{\text{voting}}$~\cite{li2023blip}
                & 62.7 & 59.1 & 61.3 & \\
                 
                BLIP-2$^{\text{concat}}$~\cite{li2023blip}
                & 62.4 & 59.7 & 60.8 & \\
                 
                SeViLa~\cite{yu2023self}
                & 63.6 & 61.3 & 61.5 & \\
                 
                \cline{1-4}\\[-2ex]
                ViperGPT+ & 64.0 & 59.8 & 67.3 & \\
                JCEF    & \underline{66.7} & \underline{61.6} & \underline{68.3} & \\
                MoReVQA & \textbf{69.2} & \textbf{64.6} & \textbf{70.2} & \\
                \bottomrule
        \end{tabular}
    }
    \vspace{-1.0mm}
    \caption{\label{table:nextqa_sota_comparison_ex1} \textbf{Expanded comparison w/state-of-the-art methods on the NExT-QA dataset.} We expand the table in the main paper and report results of MoReVQA on the official dataset subsets for ``Temporal'' and ``Causal'' questions, following prior work~\cite{yu2023self,xiao2021next} (\cite{li2023blip} results reported in \cite{yu2023self}). Our model sets a new state-of-the-art across the board, indicating the effectiveness of our overall multi-stage approach.}
    \vspace{-2.0mm}
    \end{center}
\end{table}

\begin{table}[t]
    \begin{center}
    
    \scalebox{0.95}{
        \begin{tabular}{ccccc}
                \toprule
                 
                 Method & Val & Hard-C & Hard-T & Fine-tuned \\

                 \midrule

                 \deemph{Temp[ATP]~\cite{buch2022revisiting}}                              
                 & \deemph{54.3} & \deemph{43.3} & \deemph{45.3} & \multirow{3}{*}{\cmark} \\
                 
                 \deemph{VFC~\cite{momeni2023verbs}}
                 & \deemph{58.6} & \deemph{38.3} & \deemph{39.9} &   \\

                 \deemph{HGA~\cite{jiang2020reasoning}}
                 & \deemph{49.1} & \deemph{45.3} & \deemph{43.3} &   \\

                 \deemph{HiTeA~\cite{ye2023hitea}}
                 & \deemph{63.1}  & \deemph{47.8} & \deemph{48.6} & \\
                 
                 \midrule
                 \midrule
                 VFC~\cite{momeni2023verbs}
                 & 51.5 & 30.0 & 32.2 & \multirow{3}{*}{\xmark}\\
                 
                 ViperGPT~\cite{suris2023vipergpt}                
                 & 60.0 & 56.4 & 49.8 & \\\cline{1-4}\\[-2ex]

                 \ours                          & \textbf{69.2} & \textbf{63.4} & \textbf{59.6} & \\
                 \bottomrule
        \end{tabular}
    }
    \vspace{-1.0mm}
    \caption{\label{table:nextqa_sota_comparison_ex2}\textbf{NExT-QA Hard subsets.}
    For completeness, we also explicitly report results of \ours{} on the additional ATP-Hard \cite{buch2022revisiting} subsets (Hard-C and Hard-T, following prior work \cite{suris2023vipergpt}) to compare against a selection of other prior work that also report on these subsets. We observe consistent, state-of-the-art performance across all relevant subsets, with the same trends observed in Table~\ref{table:nextqa_sota_comparison_ex1}.
    }
    \end{center}
\end{table}

\noindent\textbf{Ablation study of JCEF for varying $n$.}
In this study, we explore the impact of frame sampling rates on the performance of JCEF.
Our evaluation spanned a range of frame sampling rates from 0\% to 100\%.
At a 0\% rate, JCEF predicts without accessing video content, relying exclusively on textual questions for generating predictions, thereby omitting any visual cues.
At this baseline rate, the accuracy on NExT-QA, iVQA, and EgoSchema datatsets yielded accuracy scores of 48.5, 15.0, and 41.0, respectively.
As the frame sampling rate increased, we observe that across datasets overall accuracy reaches the peak (or near it) at about 50\% sampling rate with scores of 67.2, 54.0, and 49.9 for the respective datasets.
Beyond this point, further increases in the sampling rate resulted in diminishing returns;
at a 100\% rate -- where the model uses captions from every frame -- the accuracy scores slightly adjusted to 66.7, 56.9, and 49.9.
The results suggest that the captions become highly redundant after a certain ratio, which eventually lead less accurate predictions.

\begin{table}[t]
    \begin{center}
    
    \scalebox{0.75}{
        \begin{tabular}{ccccccc}
                \toprule

                Method & Acc$_\text{@GQA}$ & mIoP & IoP$_\text{@0.5}$ & mIoU & IoU$_\text{@0.5}$ & FT\\
                \midrule
                \deemph{IGV}~\cite{li2022invariant} & \deemph{10.2} & \deemph{21.4} & \deemph{18.9} & \deemph{{14.0}} & \deemph{{9.6}} & \multirow{4}{*}{\cmark} \\
                \deemph{Temp[CLIP]}~\cite{yang2022zero} & \deemph{16.0} & \deemph{{25.7}} & \deemph{\underline{25.5}} & \deemph{12.1} & \deemph{8.9} & \\
                \deemph{FrozenBiLM}~\cite{yang2022zero} & \deemph{\underline{17.5}} & \deemph{24.2} & \deemph{{23.7}} & \deemph{9.6} & \deemph{6.1} & \\
                \deemph{*SeViLA}~\cite{yu2023self} & \deemph{{16.6}} & \deemph{\underline{29.5}} & \deemph{22.9} & \deemph{\textbf{21.7}} & \deemph{\underline{13.8}} & \\
                \midrule
                \ours & \textbf{31.0} & \textbf{44.1} & \textbf{43.0} & \underline{20.7} & \textbf{18.3} & \xmark \\ 
                \bottomrule
        \end{tabular}
    }
    \vspace{-1.0mm}
    \caption{\label{table:next_gqa}\textbf{Comparison to SOTA on the NExT-GQA dataset.} We extend our \textit{training-free} \ours{} model to the grounded QA setting, and show strong results compared with prior finetuned work.
    Prior finetuned state-of-the-art methods are reported in \cite{xiao2023can}, and * indicates explicit localization supervision during pretraining.
    }
    \vspace{-9.0mm}
    \end{center}
\end{table}

\noindent\textbf{Extensions to other tasks.}
One of the notable strengths of \ours{} lies in its modular design, which inherently supports flexibility and ease of extension.
This architecture not only facilitates the integration of advanced models but also enables the system to adapt to a wide range of video multi-modal scenarios beyond those we initially explored, \eg, grounded videoQA and paragraph captioning.

\textbf{\em (1) Temporal grounding.}
To verify generalization beyond standard videoQA, we further evaluate our method on the task of joint temporal grounding and question-answering. Table~\ref{table:next_gqa}\footnote{Please note: Table~\ref{table:next_gqa} was updated in arXiv-v2 to reflect consistent model settings with Table \ref{table:sota_comparison}.} presents our evaluation of grounded frames, which are provided in grounding stage, on the NExT-GQA dataset \cite{xiao2023can}. Our training-free model sets a new state-of-the-art in both the key grounding (mIoP) and the combined grounded accuracy (Acc@GQA) metrics, surpassing~\cite{yu2023self} which was pre-trained with localization annotations, with strong performance on other related metrics as well.

\textbf{\em (2) Long video paragraph captioning.}
Our exploration extends to the ActivityNet-Paragraphs dataset \cite{krishna2017dense,yang2023vid2seq}, which assesses the capability of systems to generate long paragraph-level coherent summarization descriptions of long video content.
We observe our \textit{training-free} \ours{} method achieves a CIDEr score of 28.2, in comparison with Vid2Seq~\cite{yang2023vid2seq} -- the previous state-of-the-art \textit{fine-tuned} model with learnt proposals -- with a score of 28.0.
We observe that our multi-stage system is able to adapt to this alternate task setting well, \eg by employing its grounding stage to find the multiple different relevant events for summarization, and its reasoning stage to ask for more targeted event information that helps to create better overall video descriptions.

Overall, we believe these results showcase our system's adaptability to various domains and related multimodal video tasks.

\begin{figure}[t]
    \begin{center}
        \includegraphics[width=0.99\linewidth]{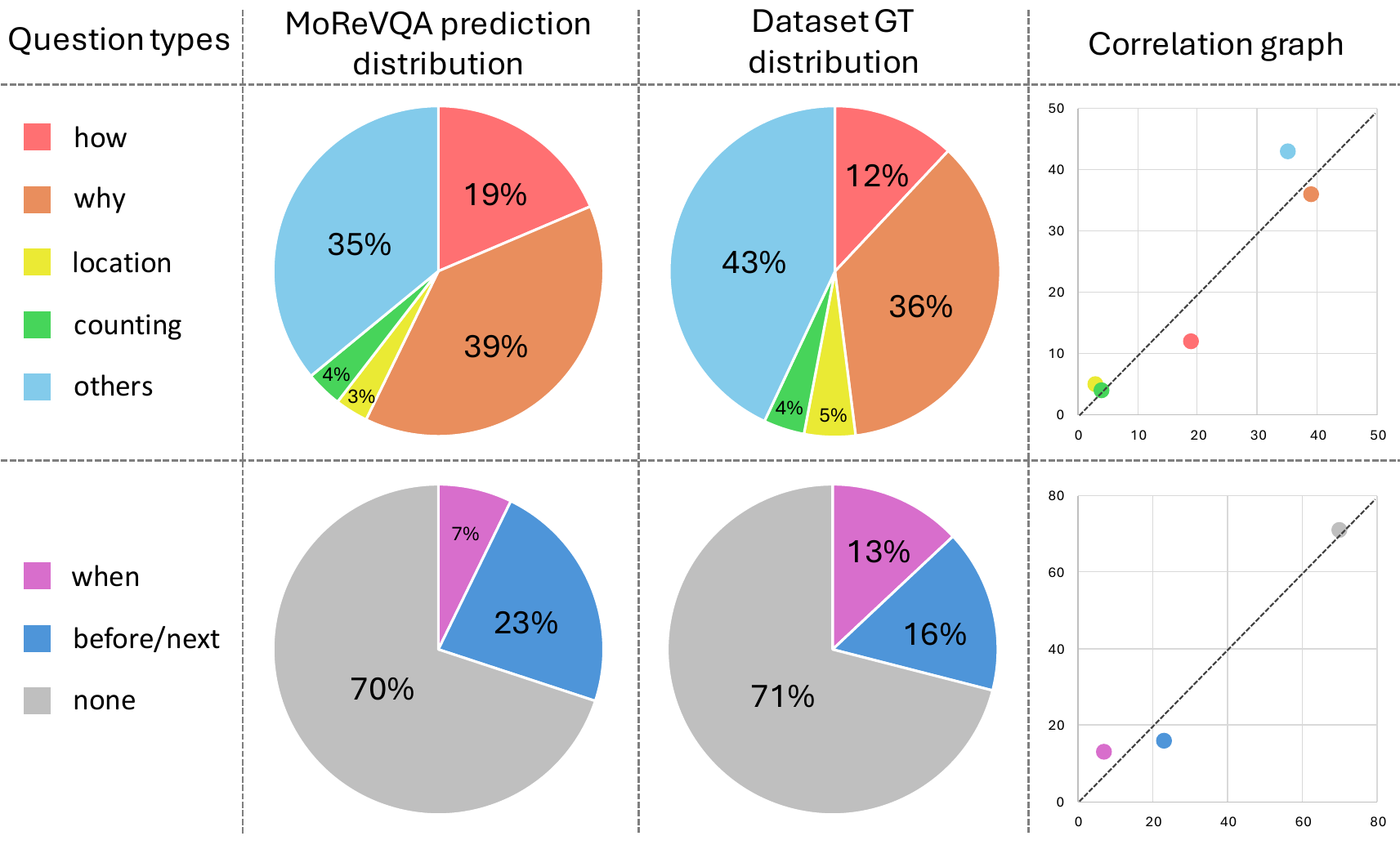}
    \end{center}
    \vspace{-1.0mm} 
    \caption{\textbf{Event parser statistics of \ours{}.}
    We observe that the event parser of our (training-free) \ours{} method naturally identifies characteristics of the underlying dataset distribution (here, NExT-QA) and this corresponds to appropriate API usage downstream in a well-correlated manner (without explicit supervision). Note that since our API is designed to work consistently across all datasets/settings (\ie the prediction values are not exactly the same as the dataset-specific metadata), we are reporting an approximate categorization of our predictions vs. the dataset-specific metadata. The graph shows the correlation between our system (x-axis) and the dataset (y-axis), where the diagonal represents an ideal 1:1 mapping.
    }
    \vspace{-1.0mm}
    \label{fig:piechart}
\end{figure}

\noindent\textbf{Usage statistics.}
To examine the effectiveness of the event parser stage -- specifically, its capacity to classify question types and identify temporal conjunctions in the questions -- we conducted a comparison between the distributions of MoReVQA (event parser) predictions and the dataset ground truth.
Figure~\ref{fig:piechart} visualizes the pie charts: the uppper two charts focus on the distribution of question type predictions (\eg, ``how'', ``why'', ``location'', ``counting'', and ``others''), while the lower two charts address the presence of temporal conjunctions within questions.
Additionally, the right-side plots illustrate the discrepancies between these two distributions.
The results suggest that our event parser stage not only proficiently classifies question types but also accurately identifies the presence of temporal conjunctions, thereby positively contributing to the subsequent stages of grounding, reasoning, as well as final predictions.

\section{Additional qualitative comparisons}
\label{sec_supp:additional_qual}
\begin{figure*}[t]
    \begin{center}
        \includegraphics[width=0.85\linewidth]{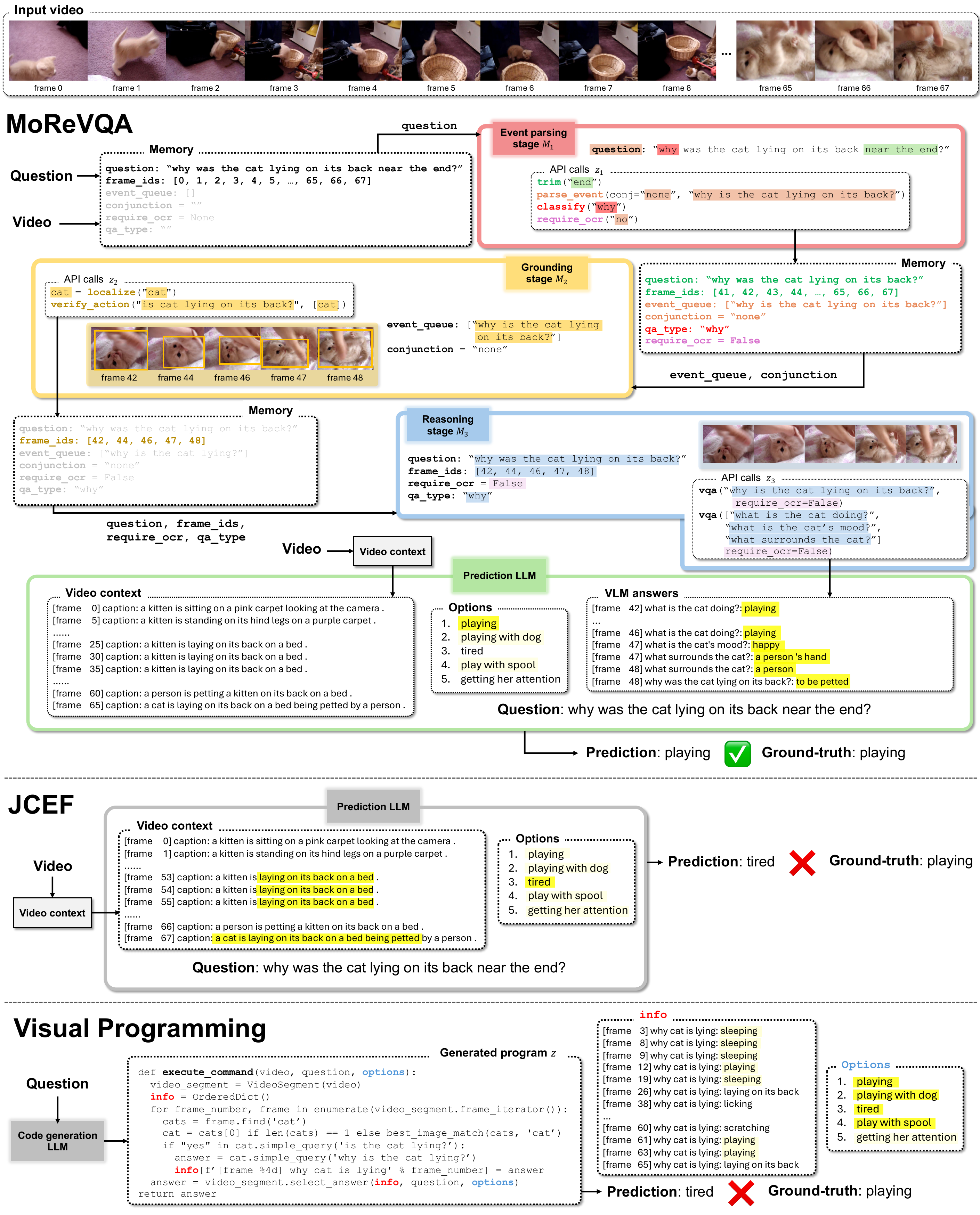}
    \end{center}
    \vspace{-5.0mm} 
\caption{\textbf{Qualitative comparison between~\ours{}, JCEF, and ViperGPT+~\cite{suris2023vipergpt}} on NExT-QA. We observe that the intermediate outputs from our \ours{} model are interpretable: event parsing stage first determines the type of question (`why'), which general area of the video should be most relevant (the `end'), parses key events from language, and other tool-use metadata. The grounding stage then determines which frames contain the `cat lying on its back', and the reasoning stage reasons about relevant sub-questions for the final answer (\eg `what is the cat doing?'), which when combined with general video-level context, gives us the final correct answer. For the same example, JCEF and ViperGPT+ obtain the wrong answer with less interpretable intermediates:
JCEF performs its best guess based solely on the general (not question-specific) captions, mostly consisting of the event `\texttt{laying...}' which leads to the prediction `\texttt{tired}'. Meanwhile, ViperGPT+ program execution shows a grounding failure of `\texttt{cat lying on its back}', returning early, irrelevant frames of 3, 8, 9, 19 (as seen in \textbf{\red{\texttt{info}}}) and implying an error \textit{how} it employs the same underlying API tools in its generated program (\eg, arguments, logic flow, etc.).
}
    \vspace{-3.0mm}
    \label{fig:supp_qual_4703}
\end{figure*}

\begin{figure*}[t]
    \begin{center}
        \includegraphics[width=0.83\linewidth]{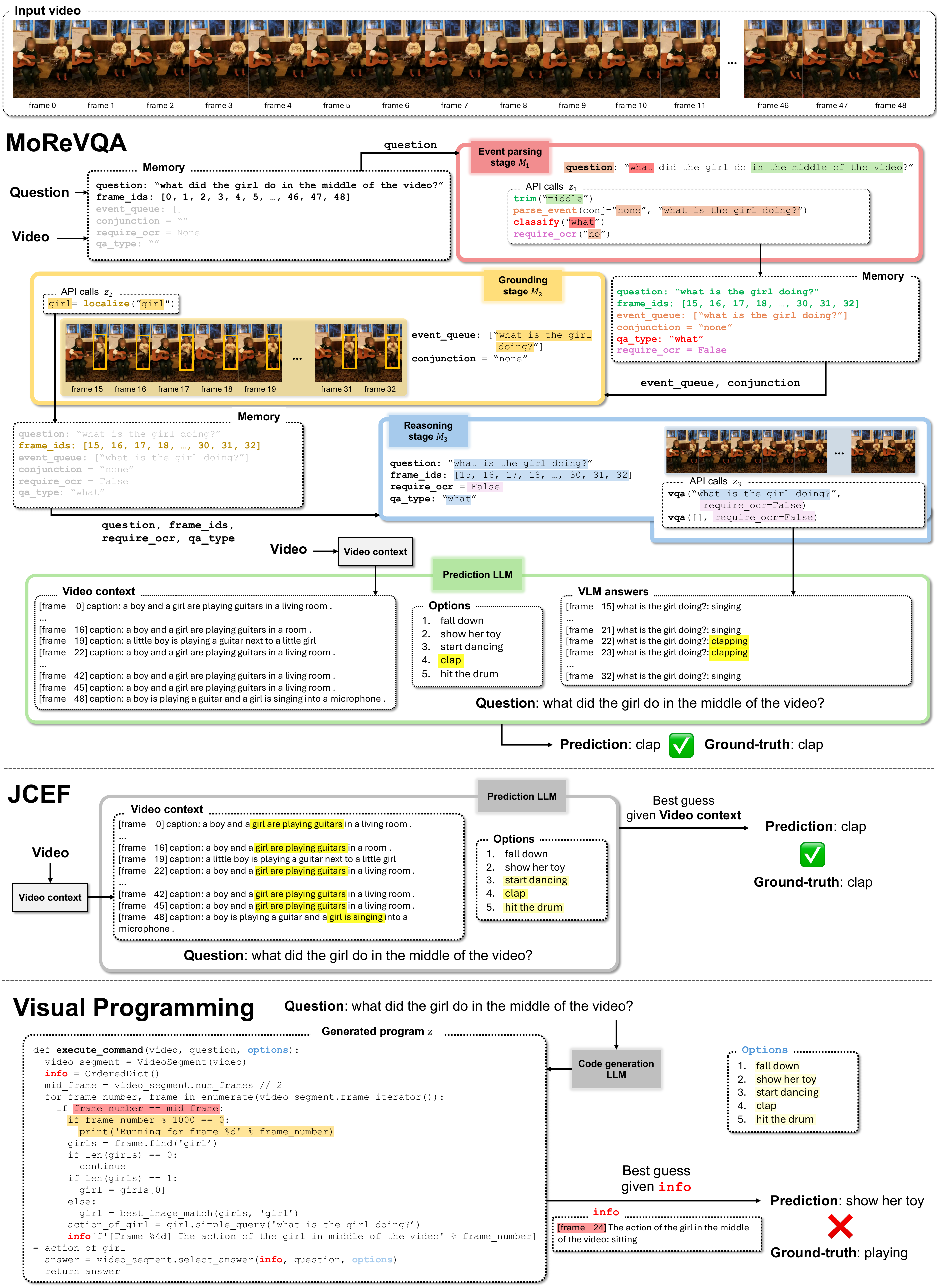}
    \end{center}
    \vspace{-5.0mm} 
    \caption{\textbf{Example qualitative comparison on NExT-QA.} In the event parsing stage of MoReVQA, the LLM determines the general part of the video that should be focused on (the `middle') and parses natural language key event for the subsequent grounding stage,`what is the girl doing'. The grounding stage collects the frames which contain the `girl', and the reasoning stage queries `what is the girl doing?' to get the candidate answers for the final prediction. For the same example, JCEF has to {\em guess} to obtain the correct final answer as the prediction LLM is not aware of if the girl is actually clapping in the video solely based on the general video information, \eg, captions of every frame. The program generated by ViperGPT+ fails in grounding correct part of the video, focusing only on a single middle frame of the video (the code highlighted in red) with some halluciations (highlighted in light orange), thereby obtaining the wrong final answer.}
    \vspace{-9.0mm}
    \label{fig:supp_qual_1074}
\end{figure*}

\begin{figure*}[t]
    \begin{center}
        \includegraphics[width=0.72\linewidth]{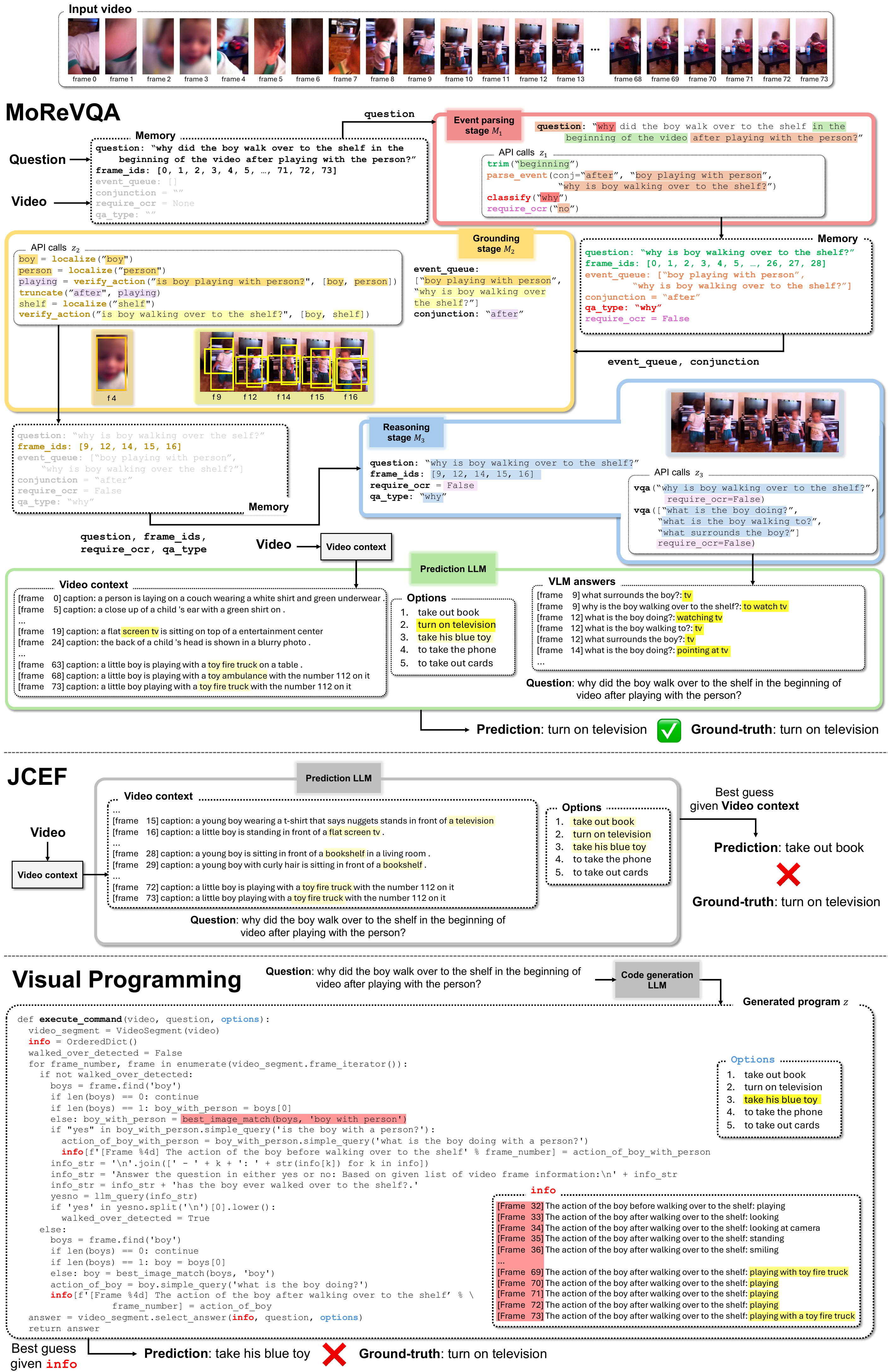}
    \end{center}
    \vspace{-5.0mm} 
    \caption{\textbf{Example qualitative comparison on NExT-QA.} The event parsing stage of MoReVQA effectively discerns general relevant part of the video (the `beginning') and parses two key events (`boy playing with person' and `why is boy walking over to the shelf') with conjunction `after'. The grounding stage first grounds the frame where the `boy is playing with person' and focuses on frames {\em after} that particular frame to localize the event `boy walking over to the shelf'. The reasoning stage reasons about relevant sub-questions for the final answer, obtaining the final correct answer.
    However, JCEF, fully relying on the general context of the video, \eg, captions, has to {\em guess} among multiple plausible candidate answers, implying that the captions alone are {\em ambiguous} and insufficient for the correct prediction.
    The program of ViperGPT+ fails in correct grounding of `boy playing with person' (highlighted in red in generated program) and `the beginning of the video' (highlighted in red in \texttt{info}), focusing on wrong temporal parts of the video and thus giving wrong final answer.
    }
    \vspace{-9.0mm}
    \label{fig:supp_qual_696}
\end{figure*}

\begin{figure*}[t]
    \begin{center}
        \includegraphics[width=0.87\linewidth]{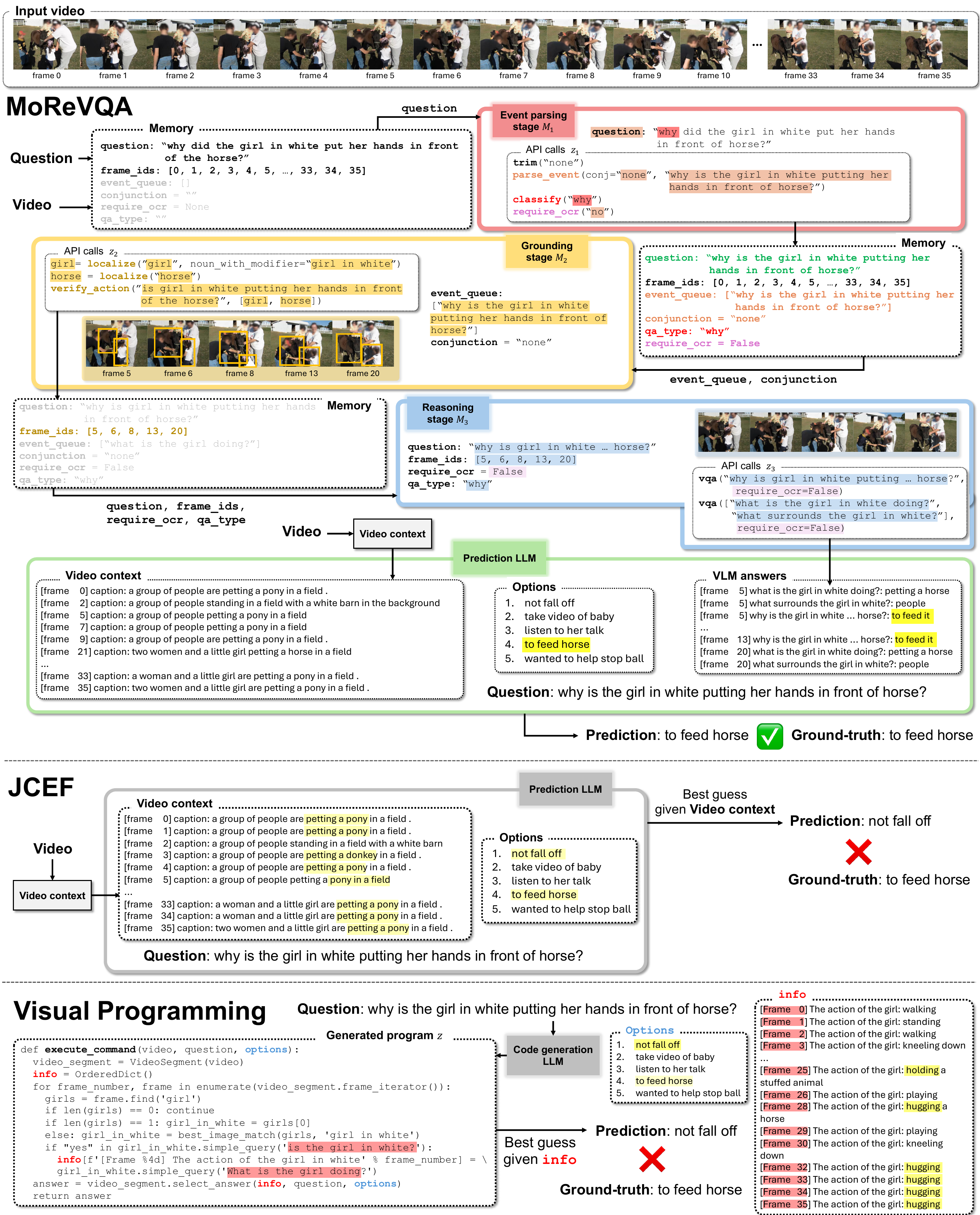}
    \end{center}
    \vspace{-5.0mm} 
    \caption{\textbf{Example qualitative comparison on NExT-QA.} Given the key event of `girl in white putting her hands in front of the horse' parsed in the event parsing stage, the grounding stage effectively grounds five most relevant frames of 5, 6, 8, 13, and 20. The reasoning stage queries only on those five pertinent frames to get the correct answer, `to feed the horse' based on VQA answers on the frames 5 and 13. Meanwhile, JCEF relying solely on the general video information, has to guess the answer based on natural language event `petting a pony', thereby giving `not fall off' as the wrong final answer. The ViperGPT+ program fails in correct grounding of `girl in white putting her hands in front of horse' and instead it grounds `girl in white' (highlighted red in generated program) and gives mostly whole part of the video (highlighted in red in \texttt{info}), making the prediction LLM difficult to predict correct answer. Also, it reasons about `what is the girl doing', not `why is the girl putting her hands in front of horse' (highlighted red in generated program), giving the wrong final answer.
    }
    \vspace{-9.0mm}
    \label{fig:supp_qual_104}
\end{figure*}

\begin{figure*}[t]
    \begin{center}
        \includegraphics[width=0.88\linewidth]{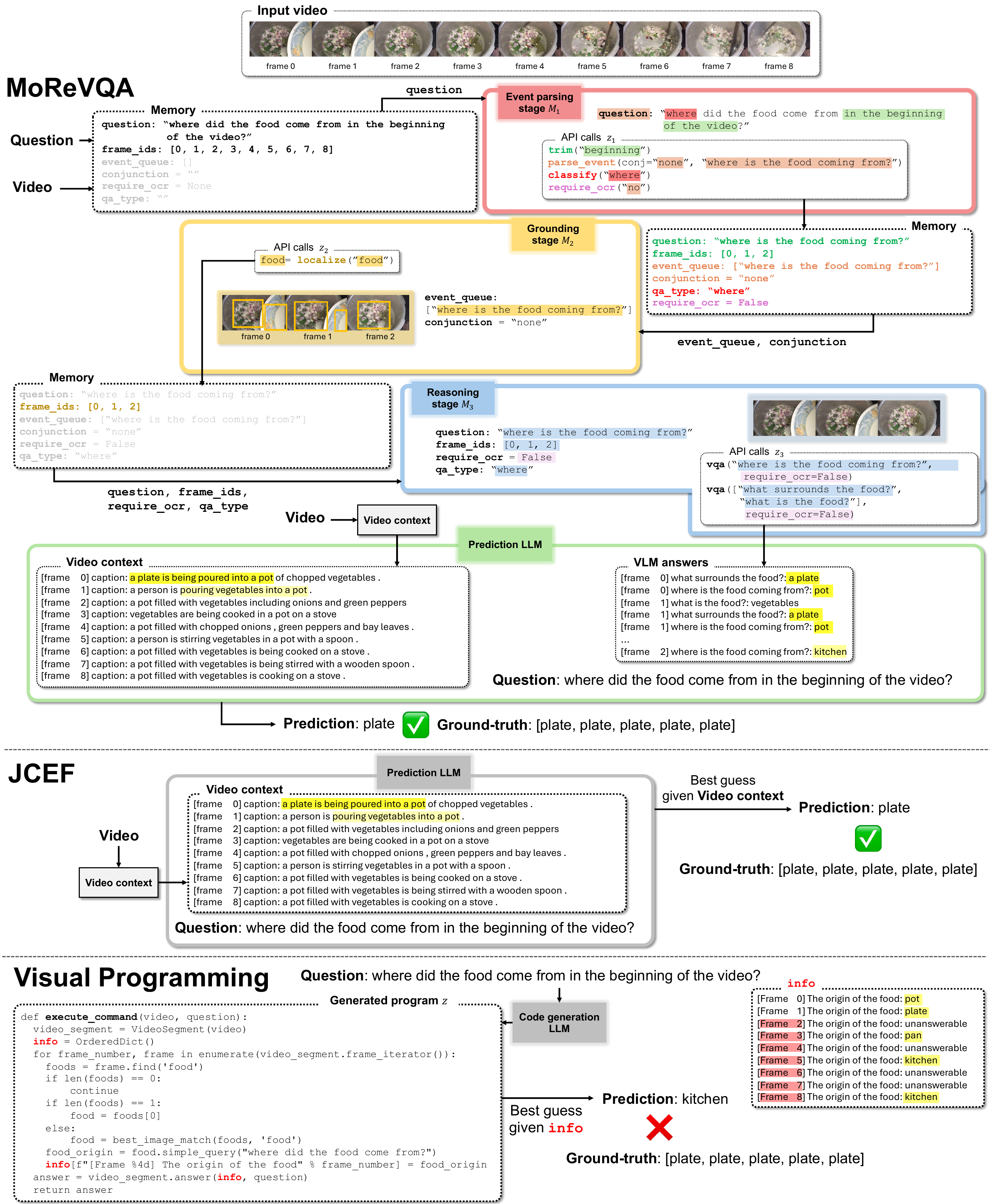}
    \end{center}
    \vspace{-5.0mm} 
    \caption{\textbf{Example qualitative comparison on iVQA.} 
    MoReVQA correctly identifies general parts of the video that should be focused on (the `beginning') in the event parsing stage, performs correct grounding of `food' in the grounding stage, and reasons about relevant sub-questions for the final answer, \eg, `what surrounds the food?'.
    JCEF also gets the correct final answer as the captions consist of relevant information to predict the final answer in the beginning of the video, \eg, \texttt{[frame 0] caption: a plate is being poured into a pot of chopped vegetables}.
    On the other hand, ViperGPT+ generates a program that localizes the `food' and return all the video frames (highlighted in red in \texttt{info}), ignoring the strong hint given in the question, `beginning'. Due to the ambiguous video information in the \texttt{info}, the program returns the wrong answer of `kitchen'.
    }
    \vspace{-9.0mm}
    \label{fig:supp_qual_i204}
\end{figure*}

\begin{figure*}[t]
    \begin{center}
        \includegraphics[width=0.85\linewidth]{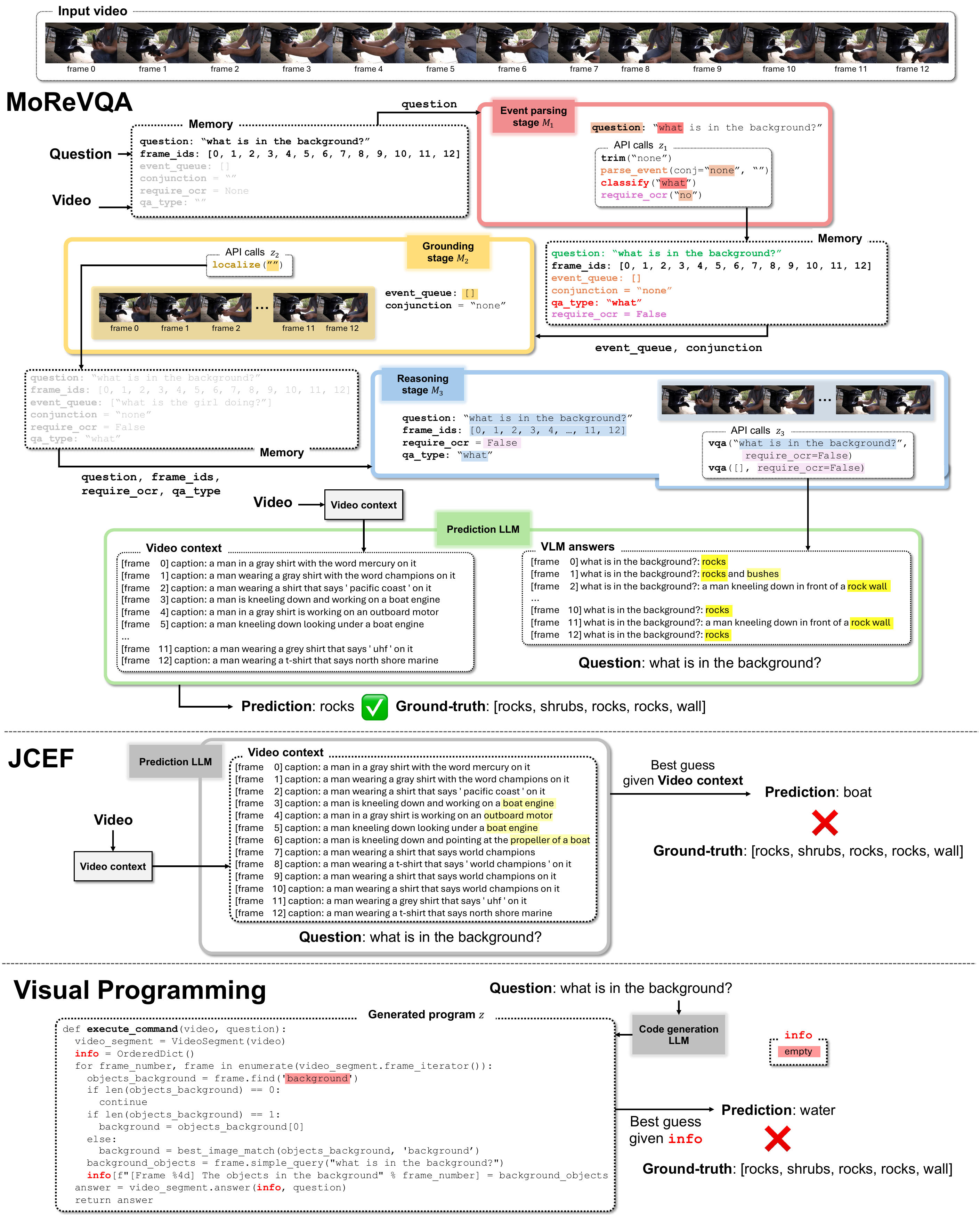}
    \end{center}
    \vspace{-5.0mm} 
    \caption{\textbf{Example qualitative comparison on iVQA.} 
    This example shows a simple question `what is in the background' for a short video of 13 frames (sampled at 1 frame per second), and highlights the capacity of \ours{} to \textit{flexibly} use its multistage pipeline in the most effective manner (it is able to ``skip'' stages implicitly when needed).
    In the event parsing stage, MoReVQA discerns the question type (`what') and at the same time, recognizing the simplicity of the input question, does not store any particular events for grounding in the \texttt{event\_queue} to effectively skip the grounding stage. Due to its simplicity and directness, the reasoning stage reasons about the input question without any supporting questions, obtaining the correct final answer `rocks'.
    With the only option of leveraging general video information in the captions, JCEF gives the wrong answer of `boat'.
    The generated program of ViperGPT+ tries to localize an object `background' (highlighted in red in generated program) which apparently is not an proper {\em object} to be localized, so it returns an empty \texttt{info} (highlighted in red in \texttt{info}). Without grounded video information, ViperGPT+ then predicts the answer solely based on the text information available in the question, giving the wrong answer `water' as the final prediction (note: counterfactually, if its output program logic been more robust, the (currently unexecuted) API call to \texttt{simple\_query} could have led to acquiring correct relevant information).
    }
    \vspace{-9.0mm}
    \label{fig:supp_qual_i101}
\end{figure*}

\begin{figure*}[t]
    \begin{center}
        \includegraphics[width=0.85\linewidth]{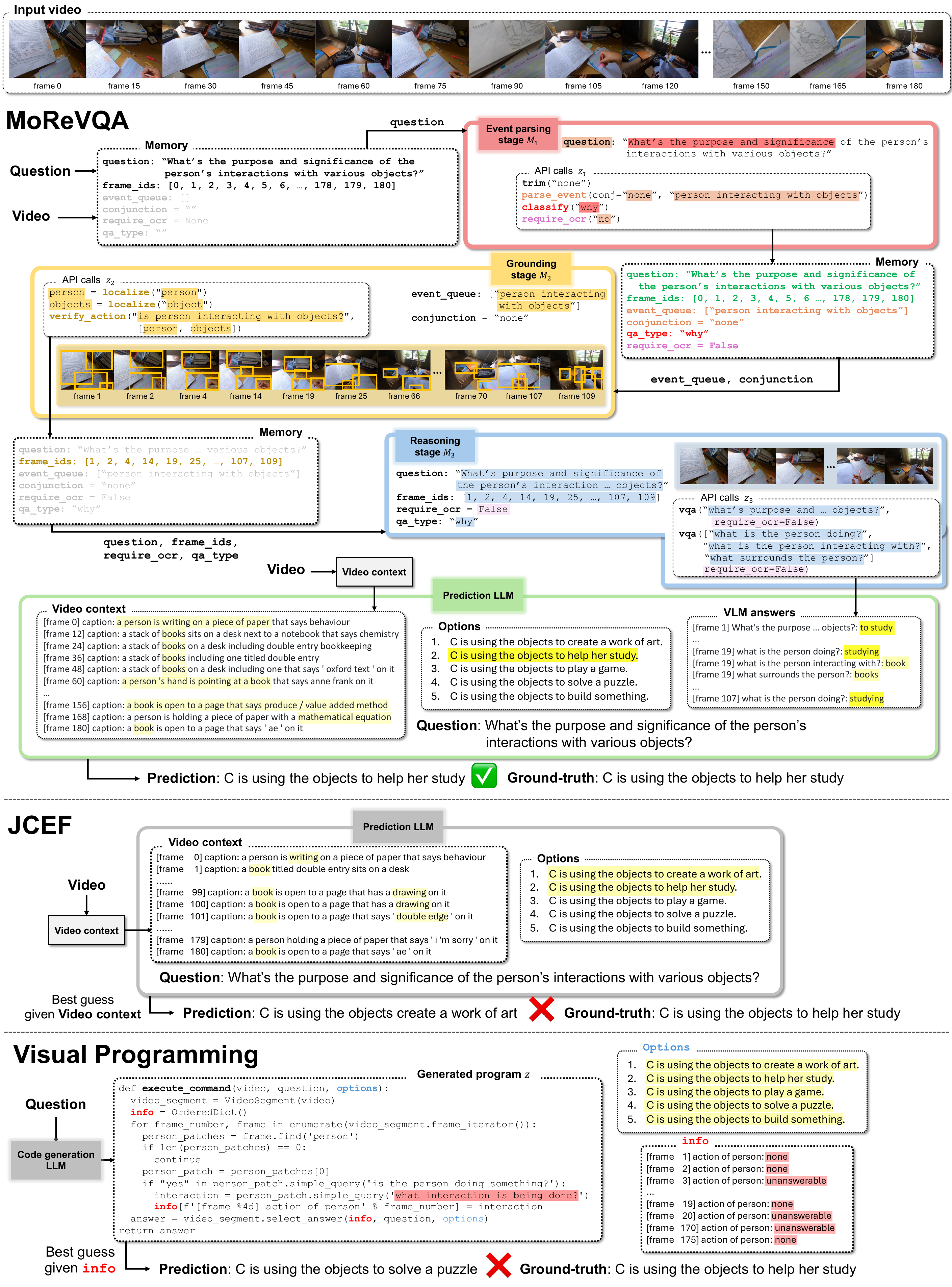}
    \end{center}
    \vspace{-5.0mm} 
    \caption{\textbf{Example qualitative comparison on EgoSchema.} MoReVQA adeptly identifies key event of `person-object interactions' during its event parsing and grounding stages. Given 'why' question type, the reasoning stage probes with supporting questions like `what is the person doing?' and `what is the person interacting with?', thus securing the correct answer. Conversely, JCEF, limited to general caption information, faces ambiguity in choosing between plausible options of `C is using objects to create a work of art' and `C is using the objects to aid her studies', as it is presented with objects like `book' and `drawing'. ViperGPT+ fails in querying exact question, instead querying `what interaction is occurring?', thus giving uninformative `\texttt{info}' for the LLM to deduce the right answer which implies the limitation of its single-stage reasoning framework in terms of generalizability.
    }
    \vspace{-9.0mm}
    \label{fig:supp_qual_e350}
\end{figure*}

In this section, we present additional qualitative analysis of our proposed method MoReVQA, comparing it with the baselines of JCEF and ViperGPT+.
Figures \ref{fig:supp_qual_4703}-\ref{fig:supp_qual_e350} demonstrate the qualitative analysis.

In our comparative analysis, we generally observe that MoReVQA consistently shows superior ability in temporal grounding and understanding question types, as seen in various scenarios on NExT-QA in Figures \ref{fig:supp_qual_4703}, \ref{fig:supp_qual_1074}, \ref{fig:supp_qual_696}, and \ref{fig:supp_qual_104}.
Particularly, the proposed MoReVQA excelled in discerning relevant parts of a video by leveraging hints in the input questions, formulating auxiliary questions to aid in prediction, and focusing on pertinent frames for accurate reasoning.
This contrasts sharply with JCEF, which often resorts to {\em guesswork}, and ViperGPT+, which frequently generates {\em brittle programs} that pose irrelevant questions or focus on incorrect parts of the video during execution, leading to inaccurate predictions. We re-iterate that the base models/APIs used in all baselines are consistent with our method; the advances in our grounding and execution are a consequence of how well our system is able to effectively utilize them in a more robust manner.

The analysis on iVQA further reinforced these observations as seen in Figures \ref{fig:supp_qual_i204} - \ref{fig:supp_qual_i101}, and demonstrate the generality of our multi-stage approach across different domains and in an ``open-ended'' question-answer setting.
MoReVQA adeptly identified relevant video parts, even in simple scenarios, and efficiently bypassed unnecessary grounding stages, directly leading to accurate predictions.
This multi-stage approach notably differed from JCEF's reliance on general video information, often resulting in incorrect answers and ViperGPT+'s inability in effective generalization or localization, thus highlighting its limitations in handling varied scenarios.
These comparative qualitative results thus underscore the robustness of the proposed MoReVQA in event understanding, accurate grounding, and effective reasoning, establishing its superiority over our simple, strong Socratic baseline (JCEF) as well as the previous state-of-the-art single-stage program generation method (ViperGPT+) in complex video question answering tasks.

We provide additional discussion in the captions accompanying each of the examples, and additional examples and visualizations are included in the \textit{full supplement} available on the project website.

\section{Broader Impacts and Future Work}
\label{sec_supp:conclusion}
In this supplement, we've described key additional details, observations, analysis, and results for our proposed MoReVQA model and related baselines. We conclude with remarks on the broader impacts, limitations, and areas for future work of our approach.

\para{Broader Impacts.} Large-scale foundation model systems have shown great potential for effectively addressing a range of tasks across many domains~\cite{singhal2023large,brohan2023rt2}, and our method demonstrates how we can build on top of, and beyond, such models \cite{zeng2022socratic,palm2,clip2021}. However, the real-world challenges regarding potential for model bias~\cite{mitchell2019model,agarwal2021evaluating} in these base models are naturally inherited by our system. As such, any system based on the work presented should have proper precautions and considerations in place.

\para{Limitations and Future Work.} Our \ours{} system presents a new path forward for modular videoQA systems, whereby we can fundamentally improve \textit{beyond} the constituent base models and module APIs in way that prior work with single-stage planning was not able to effectively capitalize on. However, there remain a number of key limitations that can be addressed as part of continual future work that we wish to discuss here: \textit{first}, like prior work in modular systems, the performance of our system is fundamentally constrained by the base modules \cite{minderer2022simple,clip2021,palm2}; as these continue to improve, we expect our overall system to also continue to improve (and importantly, to continue to \textit{outperform} the base models in a meaningful way).
\textit{Further}, our analysis has highlighted key limitations in the existing tasks and benchmarks for video understanding, particularly as they relate to complex temporal reasoning (furthering analysis from prior work \cite{buch2022revisiting,lei2022revealing,bagad2023test}); to our surprise, even recent benchmarks for long video understanding like EgoSchema \cite{mangalam2023egoschema} have significant areas for improvement to reduce key sources of bias. As later iterations of these benchmarks continue to improve towards assessing temporal, long video understanding, we believe that new opportunities for further improvements to our system will similarly emerge. \textit{Finally}, our system is focused on the challenging VideoQA task, since it emphasizes spatiotemporal reasoning across many frames of visual input and offers a challenging testbed for video-language reasoning. We do show our system generalizes well across different domains (long videos, instructional, egocentric, etc.) and videoQA types (closed set multiple choice, open vocabulary answer generation, etc.), as well as extensions of our work to settings which require language grounding (NeXT-GQA) and to video paragraph captioning (ActivityNet-Paragraphs). However, while we believe this is a strong assessment of the general potential applicability of our system, extensions to other video-language settings (and addressing any further limitations there) remains an exciting area of future work.

\clearpage

\clearpage

\end{document}